\documentclass{article}

\usepackage{microtype}
\usepackage{graphicx}
\usepackage{subfigure}
\usepackage{booktabs} 
\usepackage{multirow}
\usepackage{tikz}
\usepackage{threeparttable}
\usepackage{tikz-cd,mathtools}
\usepackage{mathtools}
\usetikzlibrary{arrows, matrix} 
\usepackage{hyperref}

\usepackage[accepted]{icml2023}

\usepackage{amsmath}
\usepackage{amssymb}
\usepackage{mathtools}
\usepackage{amsthm}

\usepackage[capitalize,noabbrev]{cleveref}

\theoremstyle{plain}
\newtheorem{theorem}{Theorem}[section]

\newtheorem{lemma}[theorem]{Lemma}

\theoremstyle{definition}

\newtheorem{assumption}[theorem]{Assumption}
\theoremstyle{remark}
\newtheorem{remark}[theorem]{Remark}
\definecolor{darkblue}{rgb}{0.0, 0.0, 0.55}
\definecolor{darkred}{rgb}{0.55, 0.0, 0.0}

\usepackage[textsize=tiny]{todonotes}

\icmltitlerunning{Solving High-Dimensional PDEs with Latent Spectral Models}

\begin{document}

\twocolumn[
\icmltitle{Solving High-Dimensional PDEs with Latent Spectral Models}

\begin{icmlauthorlist}
\icmlauthor{Haixu Wu}{software}
\icmlauthor{Tengge Hu}{software}
\icmlauthor{Huakun Luo}{software}
\icmlauthor{Jianmin Wang}{software}
\icmlauthor{Mingsheng Long}{software}
\end{icmlauthorlist}

\icmlaffiliation{software}{School of Software, BNRist, Tsinghua University.
\\ Haixu Wu $<$whx20@mails.tsinghua.edu.cn$>$}
\icmlcorrespondingauthor{Mingsheng Long}{mingsheng@tsinghua.edu.cn}

\icmlkeywords{Operator Learning}

\vskip 0.3in
]

\printAffiliationsAndNotice{}

\begin{abstract}
Deep models have achieved impressive progress in solving partial differential equations (PDEs). 
A burgeoning paradigm is learning neural operators to approximate the input-output mappings of PDEs. 
While previous deep models have explored the multiscale architectures and various operator designs, they are limited to learning the operators as a whole in the coordinate space.
In real physical science problems, PDEs are complex coupled equations with numerical solvers relying on discretization into high-dimensional coordinate space, which cannot be precisely approximated by a single operator nor efficiently learned due to the curse of dimensionality.
We present Latent Spectral Models (LSM) toward an efficient and precise solver for high-dimensional PDEs. 
Going beyond the coordinate space, LSM enables an attention-based \emph{hierarchical projection network} to reduce the high-dimensional data into a compact latent space in linear time. 
Inspired by classical spectral methods in numerical analysis, we design a \emph{neural spectral block} to solve PDEs in the latent space that approximates complex input-output mappings via learning multiple basis operators, enjoying nice theoretical guarantees for convergence and approximation. 
Experimentally, LSM achieves consistent state-of-the-art and yields a relative gain of 11.5\% averaged on seven benchmarks covering both solid and fluid physics.
Code is available at \href{https://github.com/thuml/Latent-Spectral-Models}{https://github.com/thuml/Latent-Spectral-Models}.
\end{abstract}

\section{Introduction}
Extensive real-world phenomena are governed by underlying partial differential equations (PDEs), such as turbulence, atmospheric circulation and stress of deformed materials \cite{Wazwaz2002PartialDE,roubivcek2013nonlinear}. Thus, solving PDEs is the shared foundation problem among many scientific and engineering areas and can further benefit essential real-world applications, like airflow modeling for airfoil design, atmospheric simulation for weather forecasting and stress test in civil engineering. Recently, deep models have achieved great progress in various tasks \cite{He2016DeepRL,Devlin2019BERTPO,liu2021Swin}. In view of the great nonlinear modeling capability of deep models, they have been widely used to solve PDEs by approximating the mapping between input-output pairs in PDE-governed tasks \cite{Hao2022PhysicsInformedML,Raissi2019PhysicsinformedNN,lu2021learning,li2021fourier,Cao2021ChooseAT}.

\begin{figure*}[t]
\begin{center}
    \centerline{\includegraphics[width=1.0\textwidth]{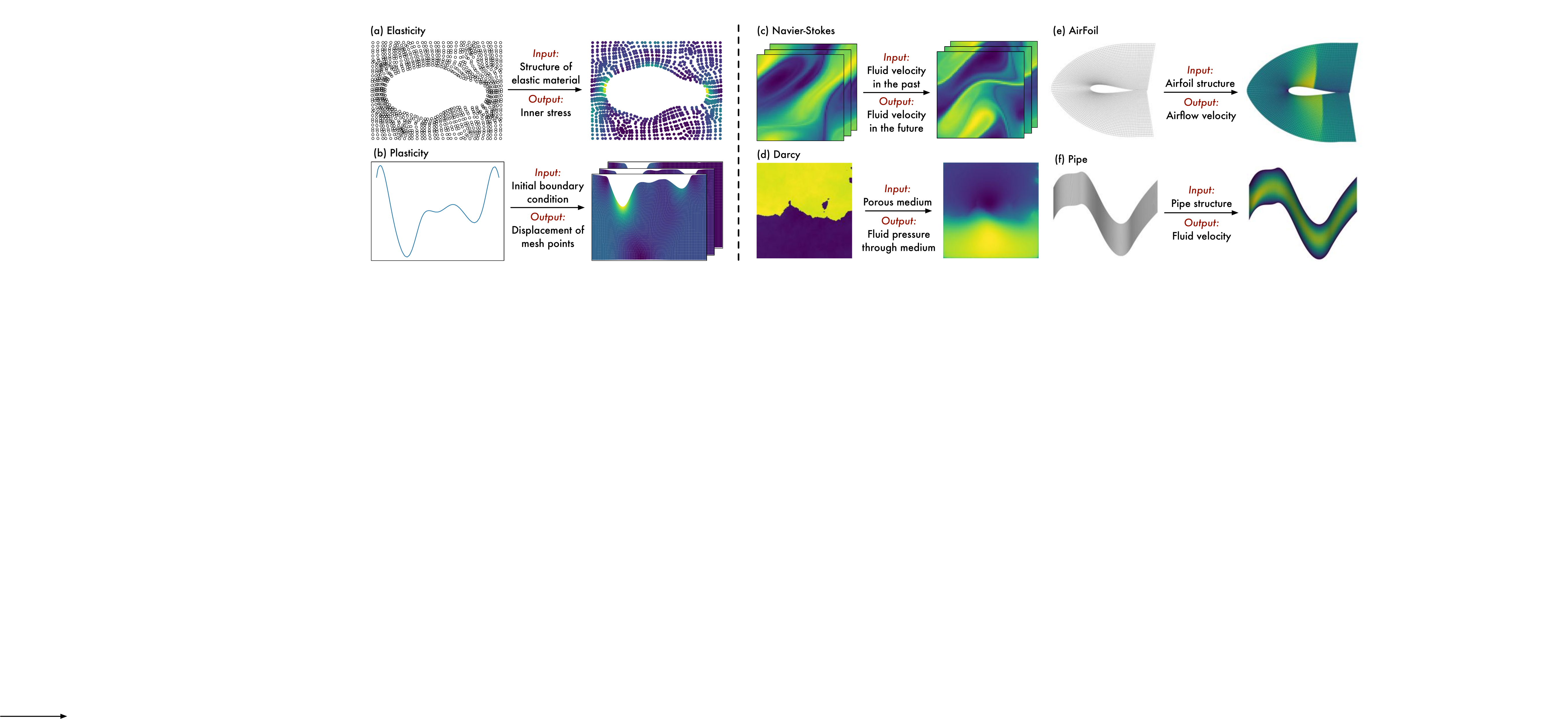}}
    \vspace{-5pt}
	\caption{Examples of PDE-governed tasks, including solid (left) and fluid (right) physics, whose solving processes are approximating complex input-output mappings in discretized high-dimensional coordinate spaces. All the tasks are covered in our experiments.}
	\label{fig:example}
\end{center}
\vspace{-20pt}
\end{figure*}

Concretely, in real-world applications, PDEs are usually discretized into high-dimensional coordinate spaces, such as point cloud, mesh and grid. For example, as shown in Figure~\ref{fig:example}(c), the fluid simulation task governed by spatiotemporal continuous Navier-Stokes equations \cite{temam2001navier} can be discretized into successive grid frames, where the dimension of coordinate space is equal to the number of pixels in all frames. However, this high-dimensionality will bring thorny challenges to the solving process. Firstly, according to the phenomenon of curse of dimensionality \cite{Trunk1979APO,Han2017SolvingHP}, the solving process will cause huge computation cost in the high-dimensional space. Secondly, due to intricate interactions among multiple physical variates of coupled equations in high-dimensional coordinate space, the input-output mappings will be too complex to be approximated by a rough deep model \cite{Trunk1979APO,karniadakis2021physics}. Thus, \emph{how to efficiently and precisely approximate complex mappings between high-dimensional input-output pairs} is the key problem to solving PDEs.

In previous works, the well-acknowledged paradigm is to learn neural operators to approximate the complex input-output mappings \cite{Li2020NeuralOG,lu2021learning}. Extensive designs of operators have been proposed, such as approximating the integral operator in Fourier space \cite{li2021fourier,anonymous2023factorized}, capturing the global information by Transformers \cite{NIPS2017_3f5ee243,Cao2021ChooseAT,anonymous2023htnet} and etc. Note that all these designs attempt to learn the operator as a whole to approximate input-output mappings. However, in high-dimensional space, the input-output mappings can be too complex to be covered by a single operator, which may suffer from optimization problems and limited performance. Besides, some works introduce the multiscale architecture into deep models \cite{Wen2021UFNOA,rahman2022u}. Although they can downsample the input into various scales, they are still limited to learning operators in the coordinate space, thereby still undergoing high-dimensionality challenges to some extent.

To tackle the above challenges, we start from the inherent property of PDE-governed tasks. It is observed that all their inputs and outputs follow certain PDE constraints, indicating that these high-dimensional data can be projected to a more compact latent space. Based on this insight, we propose the Latent Spectral Models (LSM) with a \emph{hierarchical projection network}. Different from solely downsampling data like previous methods, by leveraging the attention mechanism with latent tokens as physical prompts, our projection network can reduce the high-dimensional data into compact latent space in linear time, which will also highlight the physics properties and remove the redundant coordinate information. Benefiting from this projection, LSM can get rid of the unwieldy coordinate space and solve PDEs in the latent space. Besides, to tackle the complex mappings, inspired by the classical spectral methods in numerical analysis \cite{gottlieb1977numerical}, we present the \emph{neural spectral block} to decompose complex nonlinear mappings into multiple basis operators, which also holds the universal approximation capacity with theoretical guarantees. Experimentally, LSM achieves consistent state-of-the-art on seven well-established benchmarks and also presents good transferability between PDEs of different conditions. Our contributions are summarized as follows:
\begin{itemize}
\vspace{-5pt}
  \item Instead of solving PDEs in the coordinate space, we present the LSM with a hierarchical projection network, which can reduce high-dimensional data into compact latent space with linear complexity.
  \item Inspired by spectral methods, we propose the \emph{neural spectral block} to tackle complex mappings by learning multiple basis operators, which holds the universal approximation capacity under theoretical guarantees.
  \item LSM achieves an 11.5\% relative error reduction with respect to the previous state-of-the-art models averaged from seven PDE-solving benchmarks, covering representative PDEs in both solid and fluid physics, and also presents favorable efficiency and transferability.
\end{itemize}

\section{Preliminaries}
\subsection{Spectral Methods}\label{sec:spec}
Spectral methods are widely acknowledged in applied mathematics and scientific computing in solving PDEs numerically \cite{gottlieb1977numerical,fornberg1998practical,kopriva2009implementing}. The key idea is to approximate the solution $f$ of a certain PDE as a finite sum of $N$ orthogonal basis functions $\{f_1, f_2, \cdots, f_{N}\}$. Concretely, the approximation solution $f^{N}$ can be formulized as follows:
\begin{equation}
	\begin{split}\label{equ:spectral}
		f\approx f^{N}=\sum_{i=1}^{N}w_{i}f_{i},\\
	\end{split}
\end{equation}
where $N$ is the hyperparameter and $w_{i}$ is the coefficient for $f_{i},i\in\{1,\cdots,N\}$. With the above approximation, the solving process can be simplified as optimizing coefficients $\{w_{1},w_{2},\cdots,w_{N}\}$ to make $f^{N}$ satisfy the PDE better. The spectral methods hold nice approximation and convergence properties in solving PDEs \cite{gottlieb1977numerical}.

\subsection{Deep Models for PDEs}
Due to the immense importance in extensive scientific and engineering areas, solving PDEs has attached great interest. Since it is usually impossible to work out explicit formulas for PDE solutions, many numerical methods have been explored \cite{solin2005partial,grossmann2007numerical}. However, these classical methods need to recalculate for different instances, such as different initial velocity fields in fluid simulation or different meshes in solid stress estimation. Besides, these classical methods also suffer from poor computation efficiency, especially in processing the high-dimensional data. Recently, various deep models have been developed. The mainstream works can be roughly categorized into equation-constraint and operator-learning methods.

\paragraph{Equation-constraint methods.} This category of works directly parameterizes the PDE solution as a deep model and formalizes equation constraints, e.g.~the PDEs and their corresponding initial and boundary conditions,
as the objective function \cite{Weinan2017TheDR,Raissi2019PhysicsinformedNN,Wang2020UnderstandingAM,Wang2020WhenAW}. By doing this, they can directly obtain the solution for a certain PDE through model optimization. However, these methods require the exact formalization of underlying PDEs, which is hard to acquire in real-world applications. Thus, instead of the equation-constraint methods, this paper focuses on the operator-learning paradigm, which does not need explicit PDE formalizations.

\vspace{-5pt}
\paragraph{Operator-learning methods.} This paradigm attempts to present deep models with novel architectures to approximate the mapping between input-output pairs, such as from past observations of fluid velocity to future prediction or from the structure of elastic material to inner stress. Technically, by rewriting inputs and outputs as functions w.r.t.~coordinates, the solving process can be formulized as learning operators between input-output Banach spaces. 

Some previous works have presented various designs for operators. \citeauthor{lu2021learning}~present the DeepONet as a branch-trunk architecture derived from the universal approximation theorem \cite{Chen1995UniversalAT}. FNO \cite{li2021fourier} adopts the linear transformation in the Fourier domain to approximate the integral operator. Further, geo-FNO \cite{Li2022FourierNO} is proposed to handle tasks with complex geometrics (e.g. point cloud) by transforming the data into and back from a latent uniform mesh. F-FNO \cite{anonymous2023factorized} improves FNO with the separable Fourier transform and residual connection. KNO \cite{anonymous2023koopman} enhances the temporal dynamic modeling of FNO based on the Koopman theory \cite{Brunton2021ModernKT}. Besides, MWT \cite{Gupta2021MultiwaveletbasedOL} introduces the multiwavelet-based operator, which can capture complex dependencies at various scales. SNO \cite{Fanaskov2022SpectralNO} reformulates the input and output functions as coefficients of basis functions and adopts the neural network to learn the mapping between coefficients. Recently, \citeauthor{Cao2021ChooseAT} explored the self-attention mechanism \cite{NIPS2017_3f5ee243} and presented a Galerkin-type attention with linear complexity for solving PDEs. Unlike previous methods, instead of approximating mappings with a single operator, LSM decomposes the complex nonlinear operator into several basis operators by the neural spectral block, thereby benefiting complex PDEs solving.

Other works attempt to enhance deep models with the multiscale architecture. U-FNO \cite{Wen2021UFNOA} and U-NO \cite{rahman2022u} integrate U-Net \cite{ronneberger2015u} and FNO to empower the model with multiscale processing capability. HT-Net \cite{anonymous2023htnet} incorporates the advanced Transformers \cite{NIPS2017_3f5ee243,liu2021Swin} into a hierarchical framework to capture high-frequency components in PDEs. In contrast to previous methods, LSM presents an attention-based hierarchical projection network to project high-dimensional data into compact latent space, which is free from the redundant coordinate space and focuses on the essential physical information.

\begin{figure*}[t]
\begin{center}
    \centerline{\includegraphics[width=1.0\textwidth]{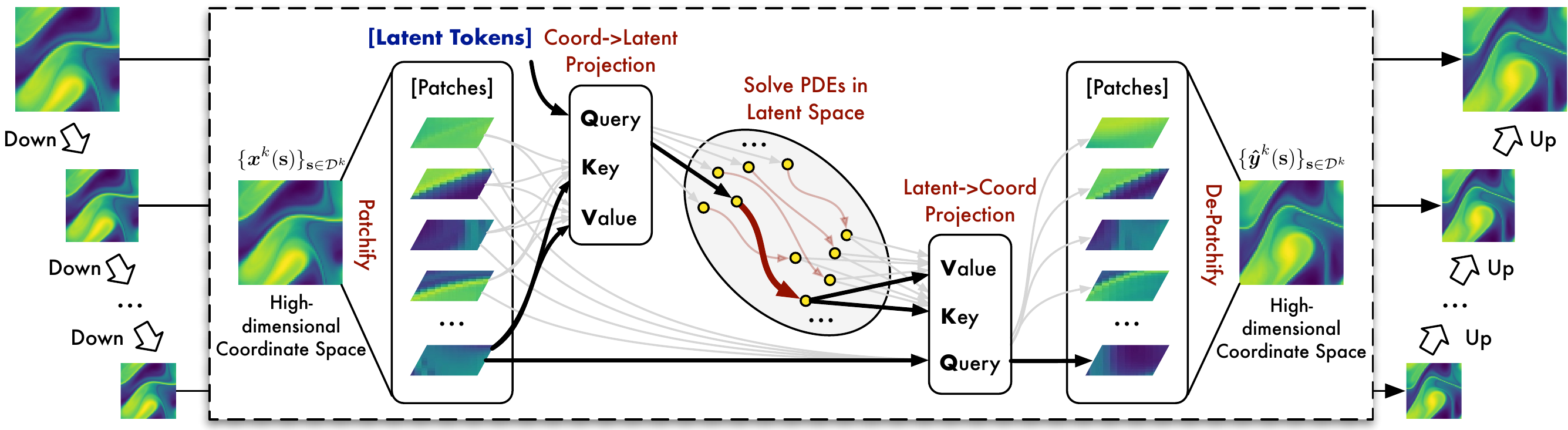}}
    \vspace{-5pt}
	\caption{Overview of LSM. The solving process is applied to each patch of each scale with three successive modules: projecting coordinate space into latent space ($\operatorname{CoordToLatent}$), solving PDEs in latent space and projecting back to coordinate space ($\operatorname{LatentToCoord}$).}
	\label{fig:framework}
\end{center}
\vspace{-20pt}
\end{figure*}

\section{Latent Spectral Models}
As aforementioned, we highlight the difficulties of solving high-dimensional PDEs as huge computation costs and complex input-output mappings. To tackle these challenges, we present LSM with a \emph{hierarchical projection network} to project the high-dimensional data into compact latent space with favorable efficiency. Further, inspired by spectral methods, we design the \emph{neural spectral block} to approximate complex mappings with multiple basis operators, which holds nice approximation and convergence properties.

\vspace{-5pt}
\paragraph{Problem setup.} For a PDE-governed task, given the coordinates in a bounded open set $\mathcal{D}\subset \mathbb{R}^{d}$, both inputs and outputs can be rewritten as functions w.r.t.~coordinates, which are in the Banach spaces $\mathcal{X}=\mathcal{X}(\mathcal{D};\mathbb{R}^{d_{\boldsymbol{x}}})$ and $\mathcal{Y}=\mathcal{Y}(\mathcal{D};\mathbb{R}^{d_{\boldsymbol{y}}})$ respectively \cite{lu2021learning,li2021fourier}. $\mathbb{R}^{d_{\boldsymbol{x}}}$ and $\mathbb{R}^{d_{\boldsymbol{y}}}$ are the range of input and output functions. For example, as Figure \ref{fig:framework} shows, both inputs and outputs are in the regular grid. Thus, $\mathcal{D}$ is a finite set of grid points within the rectangle area in $\mathbb{R}^2$. For each coordinate $\mathbf{s}\in \mathcal{D}$, $\boldsymbol{x}(\mathbf{s})\in\mathbb{R}^{d_{\boldsymbol{x}}}$ and $\boldsymbol{y}(\mathbf{s})\in\mathbb{R}^{d_{\boldsymbol{y}}}$ represent the input and output function values at position $\mathbf{s}$, corresponding to pixel values in the case of Figure \ref{fig:framework}. With the above formalization, the solving process is to approximate the optimal operator $\mathcal{F}:\mathcal{X}\to\mathcal{Y}$ with deep model $\mathcal{F}_{\theta}$, which is learned from observed samples $\{(\boldsymbol{x},\boldsymbol{y})\}$ and $\theta\in\Theta$ is the parameter set.

\vspace{-5pt}
\paragraph{Overall framework.} Instead of directly solving PDEs in high-dimensional coordinate space like previous methods, by introducing latent space, LSM can get rid of redundant coordinate information. As shown in Figure \ref{fig:framework}, LSM breaks the PDE solving process into three modules as follows:
\begin{equation}
    \begin{split}\label{equ:decompose}
\mathcal{F}_{\theta}=\mathcal{F}_{\theta_{\text{LatentToCoord}}}\circ \mathcal{F}_{\theta_{\text{Solve}}}\circ \mathcal{F}_{\theta_{\text{CoordToLatent}}}, \\
    \end{split}
\end{equation}
where $\circ$ denotes the operator composition. In LSM, the hierarchical projection network provides an attention-based instantiation for $\mathcal{F}_{\theta_{\text{CoordToLatent}}}:\mathcal{X}\to\mathcal{T}_{\mathcal{X}}$ and $\mathcal{F}_{\theta_{\text{LatentToCoord}}}:\mathcal{T}_{\mathcal{Y}}\to\mathcal{Y}$, where $\mathcal{T}_{\mathcal{X}}(\mathcal{D};\mathbb{R}^{d_{\text{latent}}})$ and $\mathcal{T}_{\mathcal{Y}}(\mathcal{D};\mathbb{R}^{d_{\text{latent}}})$ are the latent input-output Banach spaces respectively. And the neural spectral block instantiates $\mathcal{F}_{\theta_{\text{Solve}}}:\mathcal{T}_{\mathcal{X}}\to\mathcal{T}_{\mathcal{Y}}$ to approximate complex nonlinear mappings in the latent space.

\subsection{Hierarchical Projection Network}
To make the solving process free from unwieldy coordinate space, we present the hierarchical projection network by embedding attention-based projectors in a patchified multiscale architecture, which can reduce high-dimensional data into compact latent space for efficient PDE solving.

\vspace{-5pt}
\paragraph{Attention-based projectors.} If we directly apply self-attention \cite{NIPS2017_3f5ee243} among observations at multiple coordinates, the results will still be in high-dimensional coordinate space. Thus, to extract essential physical information of PDEs from redundant high-dimensional data, we propose attention-based projectors with latent tokens. The latent tokens are shared among all input-output pairs, initialized as learnable model parameters, and optimized to cover the common properties of data, namely PDE constraints, thereby providing physical prompts for projection.

Concretely, given the coordinates set $\mathcal{D}$ and the deep representations of inputs $\{\boldsymbol{x}(\mathbf{s})\}_{\mathbf{s}\in \mathcal{D}},\boldsymbol{x}(\mathbf{s})\in\mathbb{R}^{1\times d_{\text{model}}}$, we will randomly initialize $C$ latent tokens $\{\mathbf{T}_{i}\}_{i=1}^{C},\mathbf{T}_{i}\in\mathbb{R}^{1\times d_{\text{latent}}}$ to provide physical prompts. As shown in Figure \ref{fig:framework}, we adopt the latent tokens as queries and deep representations as keys and values in the attention mechanism. The residual connection is also used to ease model optimization \cite{He2016DeepRL}. This process can be formulized as:
\begin{equation}
    \begin{split}\label{equ:rawtolatent}
    \mathbf{T}_{\boldsymbol{x},i}=
    \mathbf{T}_{i}+\sum_{\mathbf{s}\in \mathcal{D}}\frac{\operatorname{Sim}\big(\mathbf{T}_{i},\boldsymbol{x}(\mathbf{s})\mathbf{W}_{\text{K}}\big)}{\sum_{\mathbf{s^\prime}\in \mathcal{D}}\operatorname{Sim}\big(\mathbf{T}_{i},\boldsymbol{x}(\mathbf{s^\prime})\mathbf{W}_{\text{K}}\big)}\left(\boldsymbol{x}(\mathbf{s})\mathbf{W}_{\text{V}}\right),
    \end{split}
\end{equation}
where $i\in\{1,\cdots,C\}$ and $\mathbf{W}_{\text{K}},\mathbf{W}_{\text{V}}\in\mathbb{R}^{d_{\text{model}}\times d_{\text{latent}}}$ are linear layers for keys and values. $\operatorname{Sim}\big(\mathbf{T}_{i},\boldsymbol{x}(\mathbf{s})\mathbf{W}_{\text{K}}\big)=\operatorname{exp}\big(\mathbf{T}_{i}\left(\boldsymbol{x}(\mathbf{s})\mathbf{W}_{\text{K}}\right)^{\sf T}\big)$ is for the similarity calculation. Under the physical prompts of learned latent tokens $\{\mathbf{T}_{i}\}_{i=1}^{C}$, the deep representations $\{\boldsymbol{x}(\mathbf{s})\}_{\mathbf{s}\in \mathcal{D}}$ in the high-dimensional coordinate space are projected to $C$ tokens $\{\mathbf{T}_{\boldsymbol{x},i}\}_{i=1}^{C}$ in latent space, where the latter is free from redundant coordinate information. To simplify notations, we summarize Eq.~\eqref{equ:rawtolatent} as $\{\mathbf{T}_{\boldsymbol{x},i}\}_{i=1}^{C}=\operatorname{CoordToLatent}(\{\mathbf{T}_{i}\}_{i=1}^{C},\{\boldsymbol{x}(\mathbf{s})\}_{\mathbf{s}\in \mathcal{D}})$.

After solving PDEs in latent space by the neural spectral block, the latent input tokens $\{\mathbf{T}_{\boldsymbol{x},i}\}_{i=1}^{C}$ are mapped to the latent output tokens $\{\mathbf{T}_{\boldsymbol{y},i}\}_{i=1}^{C}$. We summarize the solving process in latent space as $\{\mathbf{T}_{\boldsymbol{y},i}\}_{i=1}^{C}=\operatorname{Solve}(\{\mathbf{T}_{\boldsymbol{x},i}\}_{i=1}^{C})$. 

Finally, we need to project latent output tokens back to high-dimensional coordinate space as the final output. Similar to Eq.~\eqref{equ:rawtolatent}, by taking input representations as queries to provide coordinate information and latent output tokens as keys and values, this process can be formulized as follows:
\begin{equation}
    \begin{split}\label{equ:latenttoraw}
\boldsymbol{{\widehat{y}}}(\mathbf{s})=\boldsymbol{x}(\mathbf{s})+\sum_{i=1}^{C}\frac{\operatorname{Sim}\big(\boldsymbol{x}(\mathbf{s}),\mathbf{T}_{\boldsymbol{y},i}\mathbf{W}^\prime_{\text{K}}\big)}{\sum_{i^\prime=1}^{C}\operatorname{Sim}\big(\boldsymbol{x}(\mathbf{s}),\mathbf{T}_{\boldsymbol{y},i^\prime}\mathbf{W}^\prime_{\text{K}}\big)}(\mathbf{T}_{\boldsymbol{y},i}\mathbf{W}^\prime_{\text{V}}),
    \end{split}
\end{equation}
where $\mathbf{s}\in \mathcal{D}$ and $\mathbf{W}^\prime_{\text{K}},\mathbf{W}^\prime_{\text{V}}\in\mathbb{R}^{d_{\text{latent}}\times d_{\text{model}}}$ are linear layers for keys and values. Eq.~\eqref{equ:latenttoraw} is summarized as $\{\boldsymbol{{\widehat{y}}}(\mathbf{s})\}_{\mathbf{s}\in \mathcal{D}}=\operatorname{LatentToCoord}(\{\boldsymbol{x}(\mathbf{s})\}_{\mathbf{s}\in \mathcal{D}},\{\mathbf{T}_{\boldsymbol{y},i}\}_{i=1}^{C})$. The computation complexity of projectors in Eq.~\eqref{equ:rawtolatent} and \eqref{equ:latenttoraw} is linear w.r.t.~the size of coordinate set $\mathcal{D}$, namely $\mathcal{O}(|\mathcal{D}|)$. 

\vspace{-5pt}
\paragraph{Patchified multiscale architecture.} It is notable that PDEs always present different physical states according to the observed scales and regions \cite{karniadakis2021physics}. For example, in turbulent flow, unsteady vortices appear of many sizes, which interact with each other, leading to a very complex phenomenon \cite{morrison2013introduction}. To fit the intrinsic multiscale property and complex interactions of PDEs, we present a patchified multiscale architecture and attempt to solve PDEs in different regions and scales.

Technically, for the raw inputs in $\mathbb{R}^{d_{\boldsymbol{x}}}$, we firstly map them into deep representations $\{\boldsymbol{x}(\mathbf{s})\}_{\mathbf{s}\in \mathcal{D}},\boldsymbol{x}(\mathbf{s})\in\mathbb{R}^{1\times d_{\text{model}}}$ by the linear layer with parameters in $\mathbb{R}^{d_{\boldsymbol{x}}\times d_{\text{model}}}$. As shown in Figure \ref{fig:framework}, we employ the parameterized downsample layer to obtain deep representations $\{\{\boldsymbol{x}^{k}(\mathbf{s})\}_{\mathbf{s}\in \mathcal{D}^{k}}\}_{k=1}^{K}$ in $K$ scales by aggregating the local observations with learnable parameters, where $\boldsymbol{x}^{k}(\mathbf{s})\in\mathbb{R}^{1\times d_{\text{model}}^{k}}$ and $\{\boldsymbol{x}(\mathbf{s})\}_{\mathbf{s}\in \mathcal{D}}=\{\boldsymbol{x}^{1}(\mathbf{s})\}_{\mathbf{s}\in \mathcal{D}^{1}}$ is in the finest resolution. For the $k$-th scale, we further adopt the patchify operation \cite{dosovitskiy2021an} to split the coordinate set $\mathcal{D}^{k}$ into $P_{k}$ nonoverlapping patches $\{\mathcal{D}_{j}^{k}\}_{j=1}^{P_{k}}$ for different regions, where $\mathcal{D}_{j}^{k}\subset \mathcal{D}^{k}$ denotes coordinate set of the $j$-th patch. More details about downsample and patchify operations are in Appendix \ref{appendix:details}.

\begin{figure*}[t]
\begin{center}
    \centerline{\includegraphics[width=1.0\textwidth]{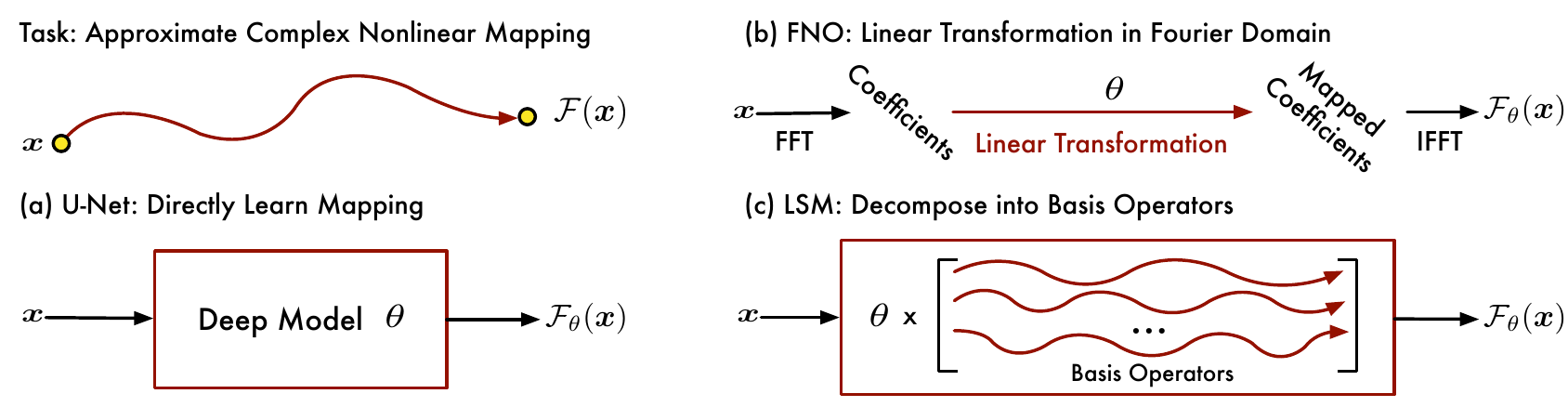}}
    \vspace{-5pt}
	\caption{Comparison in approximating complex input-output mapping. For clarity, we only keep key components for approximation.}
	\label{fig:block}
\end{center}
\vspace{-20pt}
\end{figure*}

By randomly initializing $\big\{\{\mathbf{T}_{i}^{k}\}_{i=1}^{C}\big\}_{k=1}^{K}$, $\mathbf{T}_{i}^{k}\in\mathbb{R}^{1\times d_{\text{latent}}^{k}}$ as latent tokens in $K$ scales, the solving process for the $j$-th patch in the $k$-th scale can be formulized as follows:
\begin{equation}
    \begin{split}\label{equ:overall}
\{\mathbf{T}_{\boldsymbol{x},i,j}^{k}\}_{i=1}^{C}&=\operatorname{CoordToLatent}\left(\{\mathbf{T}_{i}^{k}\}_{i=1}^{C},\{\boldsymbol{x}^{k}(\mathbf{s})\}_{\mathbf{s}\in \mathcal{D}_{j}^{k}}\right) \\
\{\mathbf{T}_{\boldsymbol{y},i,j}^{k}\}_{i=1}^{C}&=\operatorname{Solve}\left(\{\mathbf{T}_{\boldsymbol{x},i,j}^{k}\}_{i=1}^{C}\right)\\
\{\boldsymbol{{\widehat{y}}}^{k}(\mathbf{s})\}_{\mathbf{s}\in \mathcal{D}_{j}^{k}}&=\operatorname{LatentToCoord}\left(\{\boldsymbol{x}^{k}(\mathbf{s})\}_{\mathbf{s}\in \mathcal{D}_{j}^{k}},\{\mathbf{T}_{\boldsymbol{y},i,j}^{k}\}_{i=1}^{C}\right). \\
    \end{split}
\end{equation}
More details of $\operatorname{Solve}(\cdot)$ are deferred into the next section. Note that the patches in the same scale are governed by the same underlying PDEs, while in different scales, the coefficients of PDEs will change. Thus, the model parameters, e.g.~latent tokens and linear layers, are shared in patches of the same scale but independent in different scales.

After the de-patchify operation, we splice patches into the output for the $k$-th scale as $\{\boldsymbol{{\widehat{y}}}^{k}(\mathbf{s})\}_{\mathbf{s}\in \mathcal{D}^{k}}$. Then, we successively upsample the outputs in different scales from coarse to fine. Concretely, for the $k$-th scale, $\{\boldsymbol{{\widehat{y}}}^{k}(\mathbf{s})\}_{\mathbf{s}\in \mathcal{D}^{k}}$ is concatenated with the interpolation-upsampled $\{\boldsymbol{{\widehat{y}}}^{k+1}(\mathbf{s})\}_{\mathbf{s}\in \mathcal{D}^{k+1}}$ and further projected to $\mathbb{R}^{d_{\text{model}}^{k}}$ with a linear layer parameterized in $\mathbb{R}^{(d_{\text{model}}^{k+1}+d_{\text{model}}^{k})\times d_{\text{model}}^{k}}$. Finally, we obtain the finest output $\{\boldsymbol{{\widehat{y}}}(\mathbf{s})\}_{\mathbf{s}\in \mathcal{D}}$ with $\boldsymbol{{\widehat{y}}}(\mathbf{s})\in\mathbb{R}^{d_{\text{model}}}$. After the linear layer with parameters in $\mathbb{R}^{d_{\text{model}}\times d_{\boldsymbol{y}}}$, we can obtain the final output.

\subsection{Neural Spectral Block}
Benefitting from the hierarchical projection network, we can solve PDEs by approximating the complex mapping between latent input-output tokens as described in Eq.~\eqref{equ:overall}. 

As shown in Figure \ref{fig:block}, instead of learning a single operator, inspired by classical spectral methods in numerical analysis (Section \ref{sec:spec}), we present the neural spectral block by decomposing complex mappings into multiple basis operators:
\begin{equation}
	\begin{split}\label{equ:spectraloperator}
 \mathcal{F}_{\theta_{\text{Solve}}}=\sum_{i=1}^{N}w_{i}\mathcal{F}_{\theta_{\text{Solve},i}},\\
	\end{split}
\end{equation}
where $N$ is the hyperparameter and $\{\mathcal{F}_{\theta_{\text{Solve}},i}\}_{i=1}^{N}$ are orthogonal basis operators with learnable parameters $\{w_{i}\}_{i=1}^{N}$. Following the classical design in spectral methods \cite{Jackson1934TheCO,tolstov2012fourier}, we select the trigonometric basis operators. Thus, for ${\boldsymbol{t}}_{\boldsymbol{x}}:\mathcal{D}\to\mathbb{R}^{d_{\text{latent}}}\in\mathcal{{T}}_{\mathcal{X}},\forall \mathbf{s}\in\mathcal{D}$, we define the multiple basis operators as follows:
\begin{equation}
	\begin{split}\label{equ:basisdefinition}
\mathcal{F}_{\theta_{\text{Solve}},(2k-1)}\big(\boldsymbol{t}_{\boldsymbol{x}}(\mathbf{s})\big) &= \sin\big(k \boldsymbol{t}_{\boldsymbol{x}}(\mathbf{s})\big) \\
\mathcal{F}_{\theta_{\text{Solve}},(2k)}\big(\boldsymbol{t}_{\boldsymbol{x}}(\mathbf{s})\big) & = \cos\big(k \boldsymbol{t}_{\boldsymbol{x}}(\mathbf{s})\big),
	\end{split}
\end{equation}
where $k\in\{1,\cdots,\frac{N}{2}\}$ and $N$ is even. Technically, given the latent input token $\mathbf{T}_{\boldsymbol{x}}\in\mathbb{R}^{d_{\text{latent}}}$, the latent output token $\mathbf{T}_{\boldsymbol{y}}$ of the neural spectral block is calculated as follows:
\begin{equation}
	\begin{split}\label{equ:specblock}
\mathbf{T}_{\boldsymbol{y}}=\mathbf{T}_{\boldsymbol{x}}+ \mathbf{w}_{0} + \mathbf{w}_{\text{sin}}\begin{bmatrix}
\sin(\mathbf{T}_{\boldsymbol{x}})\\
\vdots \\
\sin(\frac{N}{2}\mathbf{T}_{\boldsymbol{x}})
\end{bmatrix}
+\mathbf{w}_{\text{cos}}
\begin{bmatrix}
\cos(\mathbf{T}_{\boldsymbol{x}})\\
\vdots \\
\cos(\frac{N}{2}\mathbf{T}_{\boldsymbol{x}}) \\
\end{bmatrix},
	\end{split}
\end{equation}
where $\mathbf{w}_{0}\in\mathbb{R}^{d_{\text{latent}}},\mathbf{w}_{\text{sin}}\in\mathbb{R}^{1\times {\frac{N}{2}}},\mathbf{w}_{\text{cos}}\in\mathbb{R}^{1\times \frac{N}{2}}$ are learnable parameters. Residual connection is also adopted to facilitate optimization \cite{He2016DeepRL}. We summarize the process of the neural spectral block as $\mathbf{T}_{\boldsymbol{y}}=\operatorname{Solve}(\mathbf{T}_{\boldsymbol{x}})$, which is applied to the latent input tokens of every patch at every scale. Also according to the analysis in Eq.~\eqref{equ:overall}, like latent tokens, $\mathbf{w}_{0},\mathbf{w}_{\text{sin}},\mathbf{w}_{\text{cos}}$ is shared in patches of the same scale but independent in different scales.

Since PDE constraints have already been involved in input-output pairs, during the model training, $\mathbf{w}_{0},\mathbf{w}_{\text{sin}},\mathbf{w}_{\text{cos}}$ will be optimized to satisfy the PDEs better, namely solving PDEs in latent space. Besides, the neural spectral block also holds the universal approximation capacity with favorable convergence property guaranteed by the following theorems.

\vspace{5pt}
\begin{assumption}[\textbf{Finite Coordinate Set}] 
In real-world applications, the analysis or numerical simulation of the PDE-governed task is mainly in the regular grid, mesh or point cloud, where the input is only observed on finite coordinates. Thus, to simplify the following theoretical derivations, we assume that $\mathcal{D}=\{\mathbf{s}_{1},\cdots, \mathbf{s}_{M}\}$ is a finite set with size $M$, e.g. for a frame with height $H$ and weight $W$, $M$ is $H\times W$.
\end{assumption}

\begin{remark}[\textbf{Simplification w.r.t.~Finite Coordinate Set}]
\label{remark:simplification}
By assuming that $\mathcal{D}$ is a finite set with $M$ coordinates, the learning process of operator $\mathcal{F}:\mathcal{X}(\mathcal{D};\mathbb{R}^{d_{\boldsymbol{x}}})\to\mathcal{Y}(\mathcal{D};\mathbb{R}^{d_{\boldsymbol{y}}})$ is simplified to solve the function $\boldsymbol{f}:\mathbb{R}^{M\times d_{\boldsymbol{x}}}\to\mathbb{R}^{M\times d_{\boldsymbol{y}}}$, where $\mathcal{F}(\boldsymbol{x})=\boldsymbol{f}\circ \boldsymbol{x}, \forall \boldsymbol{x}\in \mathcal{X}$. Since the channel dimension can be seen as independent, we only focus on the coordinate dimension $M$ in the following derivations.
\end{remark}

\begin{theorem}[\textbf{Convergence of Trigonometric Approximation in High-dimensional Space}]
\label{thm:universalappmulti}
\cite{dyachenko1995rate}
Let $\boldsymbol{f}:\mathbb{R}^{M}\to\mathbb{R}^{M}$ be a $2\pi$-periodic function w.r.t.~the variable on each dimension, where $\boldsymbol{f}\in L_{p}\left([-\pi,\pi)^{M}\right), M\ge 2$, $1\leq p\leq \infty$ and $p\neq 2$. For $\boldsymbol{f}$ defined on the $M$-dimension space, its trigonometric approximation $\boldsymbol{f}^{N}$ is defined as
\begin{equation}
	\begin{split}\label{equ:multipleseries}
\boldsymbol{f}^{N}(\mathbf{x})=\sum_{\mathbf{k}\in\mathbb{Z}^{M},|\mathbf{k}|\leq N}\left(\frac{1}{2\pi}\int_{{[-\pi,\pi)^{M}}}\boldsymbol{f}(\mathbf{t})e^{-i\mathbf{k}\mathbf{t}}\mathrm{d}{\mathbf{t}} \right)e^{i\mathbf{k}\mathbf{x}},
	\end{split}
\end{equation}
If $\boldsymbol{f}$ satisfies the Lipschitz condition, namely there is a non-negative constant $K_1$ such that
\begin{equation}
	\begin{split}\label{equ:condition}
\|\boldsymbol{f}(\mathbf{x})-\boldsymbol{f}(\mathbf{y})\|_{p}\leq K_1\|\mathbf{x}-\mathbf{y}\|_{p},\ \forall \mathbf{x},\mathbf{y}\in\mathbb{R}^{M},
	\end{split}
\end{equation}
and if ${(M-1)|\frac{1}{2}-\frac{1}{p}|<1}$, then there exists a constant $K_2$ such that
\begin{equation}
	\begin{split}\label{equ:convergencespeedmultiple}
\big\|\boldsymbol{f}-\boldsymbol{f}^{N}\big\|_{p}\leq K_2 N^{(M-1)|\frac{1}{2}-\frac{1}{p}|-1}.
	\end{split}
\end{equation}
\end{theorem}

\begin{remark}[\textbf{Slow Convergence Rate in High-dimensional Space}]
\label{remark:convergencespeed}
As demonstrated in Theorem \ref{thm:universalappmulti}, the convergence rate of trigonometric approximation is directly related to the dimension $M$, indicating that the spectral methods suffer from the slow convergence rate for high-dimensional space, e.g. $M=H\times W$ for a frame with height $H$ and width $W$. Actually, the convergence properties of spectral methods in high-dimensional spaces are still under explored as an open problem \cite{Brandolini2020PicksTA}. These results also support our design in solving PDEs in latent space instead of high-dimensional coordinate space.
\end{remark}

\begin{remark}[\textbf{Solving Process in Latent Space}]
After projecting the $M$-dimension data into independent latent tokens and further restricting each latent token within $[0,\pi]$ through proper normalization, the solving process in the latent space is to approximate $f:[0,\pi]\to\mathbb{R}$. 
\end{remark}

\begin{theorem}[\textbf{Approximation and Convergence Properties of Neural Spectral Block}]\label{thm:universalappsingle}
Given $f:[0,\pi]\to\mathbb{R}$, if $f$ satisfies the Lipschitz condition, there is a choice of model parameters such that the approximation $f^{N}$ defined in neural spectral block (trigonometric approximation with residual) can uniformly converge to $f$ with the speed as
\begin{equation}
	\begin{split}\label{equ:convergencespeedsingle}
|f(x)-f^{N}(x)|\leq \frac{K_3\ln N}{N}, \forall x\in[0,\pi],
	\end{split}
\end{equation}
where $K_3$ is a constant that does not depend on $f$ nor $N$.
\end{theorem}
\begin{proof}
See Appendix \ref{appendix:proof}.    
\end{proof}

\section{Experiments}
We extensively evaluate the proposed LSM on seven benchmarks, covering the typical PDEs in both solid and fluid physics and samples in various geometrics. 
\begin{table}[h]
    \vspace{-10pt}
	\caption{Summary of experiment benchmarks. }
	\label{tab:dataset_summary}
	\vskip 0.1in
	\centering
	\begin{small}
		\begin{sc}
			\renewcommand{\multirowsetup}{\centering}
			\setlength{\tabcolsep}{3pt}
			\scalebox{1}{
			\begin{tabular}{l|c|c|c}
				\toprule
			    Physics & Benchmarks & Geometry & \#Dim \\
			    \midrule
			\multirow{3}{*}{Solid} & Elasticity-P & Point Cloud & 2D \\
                  & Elasticity-G & Regular Grid & 2D \\
			     & Plasticity & Structured Mesh & 3D \\
                  \midrule
                  \multirow{4}{*}{Fluid} 
		          & Navier–Stokes & Regular Grid & 3D \\	     
                    & Darcy & Regular Grid & 2D\\
			     & Airfoil & Structured mesh & 2D \\
                  & Pipe & Structured mesh & 2D \\
				\bottomrule
			\end{tabular}}
		\end{sc}
	\end{small}
	\vspace{-10pt}
\end{table}

\vspace{-5pt}
\paragraph{Benchmarks.} As shown in Table \ref{tab:dataset_summary}, the experimental samples of seven benchmarks are recorded in various geometrics, including the regular grid, point cloud and structured mesh in the 2D or 3D space. These benchmarks are generated by different PDEs for different tasks. For clearness, we summarize the tasks of all benchmarks in Figure \ref{fig:example}. Specifically, Elasticity-G is interpolated from Elasticity-P. More details can be found in Appendix \ref{appendix:dataset}, including the governing PDEs, size of benchmarks and input-output resolutions.

\vspace{-5pt}
\paragraph{Baselines.} We compare the LSM with fourteen well-acknowledged and advanced models in all seven benchmarks, including three baselines proposed for vision tasks: U-Net \citeyearpar{ronneberger2015u}, ResNet \citeyearpar{He2016DeepRL}, Swin Transformer \citeyearpar{liu2021Swin}, and ten baselines presented for PDEs: DeepONet \citeyearpar{lu2021learning}, TF-Net \citeyearpar{Wang2019TowardsPD}, FNO \citeyearpar{li2021fourier}, U-FNO \citeyearpar{Wen2021UFNOA}, WMT \citeyearpar{Gupta2021MultiwaveletbasedOL}, Galerkin Transformer \citeyearpar{Cao2021ChooseAT}, SNO \citeyearpar{Fanaskov2022SpectralNO}, U-NO \citeyearpar{rahman2022u}, HT-Net \citeyearpar{anonymous2023htnet}, F-FNO \citeyearpar{anonymous2023factorized}, KNO \citeyearpar{anonymous2023koopman}. U-NO and HT-Net are previous state-of-the-art models in solving PDEs. Note that all the above baselines are proposed for regular grid or structured mesh. Thus, for the Elasticity-P benchmark in point cloud, we adopt the special transformation proposed by geo-FNO \citeyearpar{Li2022FourierNO} at the beginning and end of these models, which can transform irregular input domain into or back from a uniform mesh. 

\vspace{-5pt}
\paragraph{Implementation.} For fairness, all the methods are trained with L2 loss and 500 epochs, using the ADAM \cite{DBLP:journals/corr/KingmaB14} optimizer with an initial learning rate of $10^{-3}$. The batch size is set to 20. We adopt the sum of mean squared error (MSE) on each coordinate as the metric. A comprehensive description is provided in Appendix \ref{appendix:details}.

\begin{table*}[t]
	\caption{Performance comparison with fourteen baselines on all benchmarks. MSE is recorded. A smaller MSE indicates better performance. For clarity, the best result is in bold and the second best is underlined. Promotion refers to the relative error reduction w.r.t.~the second best model on each benchmark. We only compare KNO \citeyearpar{anonymous2023koopman,xiong2023koopmanlab} and TF-Net \citeyearpar{Wang2019TowardsPD} on the Navier–Stokes benchmark, since they are proposed for auto-regressive tasks in fluid simulation. In addition to the quantitative performance, we also rank the models on each benchmark. See Table \ref{tab:all_efficiency} for the performance rankings.
 }
	\label{tab:mainres}
	\vspace{-5pt}
	\vskip 0.15in
	\centering
	\begin{small}
		\begin{sc}
			\renewcommand{\multirowsetup}{\centering}
			\setlength{\tabcolsep}{6pt}
			\begin{tabular}{l|ccc|cccc}
				\toprule
                    \multirow{3}{*}{Model} & \multicolumn{3}{c}{Solid Physics$^\ast$} & \multicolumn{4}{c}{Fluid Physics$^\dagger$} \\
                    \cmidrule(lr){2-4}\cmidrule(lr){5-8}
				& Elasticity-P $\ddagger$ & Elasticity-G & Plasticity & Navier–Stokes & Darcy & Airfoil & Pipe \\
				\midrule
				U-Net \citeyearpar{ronneberger2015u} & 0.0235 & 0.0531 & 0.0051 & 0.1982 & 0.0080 & 0.0079 & 0.0065 \\
                    ResNet \citeyearpar{He2016DeepRL} & 0.0262 & 0.0843 & 0.0233 & 0.2753 & 0.0587 & 0.0391 & 0.0120 \\
                    TF-Net \citeyearpar{Wang2019TowardsPD} & / & / & / & 0.1801 & / & / & / \\
                    Swin \citeyearpar{liu2021Swin} & 0.0283 & 0.0819 & 0.0170 & 0.2248 & 0.0397 & 0.0270 & 0.0109 \\
                    DeepONet \citeyearpar{lu2021learning} & 0.0965 & 0.0900 & 0.0135 & 0.2972 & 0.0588 & 0.0385 & 0.0097 \\
                    FNO \citeyearpar{li2021fourier} & \underline{0.0229} & 0.0508 & 0.0074 & 0.1556 & 0.0108 & 0.0138 & 0.0067 \\
                    U-FNO \citeyearpar{Wen2021UFNOA} &0.0239 & 0.0480 & 0.0039 & 0.2231 & 0.0183 & 0.0269 & \underline{0.0056} \\
                    WMT \citeyearpar{Gupta2021MultiwaveletbasedOL} & 0.0359 & 0.0520 & 0.0076 & \underline{0.1541} & 0.0082 & 0.0075 & 0.0077 \\
                    Galerkin \citeyearpar{Cao2021ChooseAT} &0.0240 & 0.1681 & 0.0120 & 0.2684 & 0.0170 & 0.0118 & 0.0098\\
                    SNO \citeyearpar{Fanaskov2022SpectralNO} &0.0390 &0.0987 & 0.0070 & 0.2568 & 0.0495 & 0.0893 & 0.0294\\
                    U-NO \citeyearpar{rahman2022u} & 0.0258 & \underline{0.0469} & \underline{0.0034} & 0.1713 & 0.0113 & 0.0078 & 0.0100 \\
                    HT-Net \citeyearpar{anonymous2023htnet} &0.0372 & 0.0472 & 0.0333 & 0.1847 & 0.0079 & \underline{0.0065} & 0.0059 \\
                    F-FNO \citeyearpar{anonymous2023factorized} &0.0263 & 0.0475 & 0.0047 & 0.2322 & \underline{0.0077} & 0.0078 & 0.0070 \\
                    KNO \citeyearpar{anonymous2023koopman} & / & / & / & 0.2023 & / & / & / \\
                    \midrule
                    \textbf{LSM} & \textbf{0.0218} & \textbf{0.0408} & \textbf{0.0025} & \textbf{0.1535} & \textbf{0.0065} & \textbf{0.0059} & \textbf{0.0050} \\
                    Promotion & 4.8\% & 13.0\% & 26.5\% & 0.4\% & 15.6\% & 9.2\% & 10.7\% \\
				\bottomrule
			\end{tabular}
		\end{sc}
      \begin{tablenotes}
      \footnotesize
        \item[] $\ast$  Top 5 ranking methods of solid benchmarks: LSM (ours), U-NO \citeyearpar{rahman2022u}, U-FNO \citeyearpar{Wen2021UFNOA}, FNO \citeyearpar{li2021fourier}, F-FNO \citeyearpar{anonymous2023factorized}.
        \item[] $\dagger$  Top 5 ranking methods of fluid benchmarks: LSM (ours), HT-Net \citeyearpar{anonymous2023htnet}, WMT \citeyearpar{Gupta2021MultiwaveletbasedOL}, U-Net \citeyearpar{ronneberger2015u}, F-FNO \citeyearpar{anonymous2023factorized}.
        \item[] $\ddagger$ All the experiments in Elasticity-P adopt the special transformation from geo-FNO \citeyearpar{Li2022FourierNO} to handle the point cloud geometric. Especially, FNO \citeyearpar{li2021fourier} with the special transformation is just equivalent to geo-FNO \citeyearpar{Li2022FourierNO}.
    \end{tablenotes}
	\end{small}
    \vspace{-10pt}
\end{table*}

\subsection{Main Results}
\paragraph{Results.} As shown in Table \ref{tab:mainres}, LSM achieves consistent state-of-the-art performance on all seven benchmarks, covering both solid and fluid physics, justifying the generality of LSM on different PDEs, geometrics and dimensions. Overall, LSM averagely outperforms the previous best method on each benchmark by 11.5\%. Specifically, our method accomplishes remarkable promotions on tasks with semantically heterogeneous input and output, such as 13.0\% on Elasticity-G (0.0469$\to$0.0408), 15.6\% on Darcy (0.0077$\to$0.0065). Note that these two tasks require the model to capture complex mappings between input and output, e.g.~mapping from structure to inner stress on Elasticity-G or from the porous medium to flow on Darcy. From Table~\ref{tab:mainres}, we can find that the well-acknowledged FNO performs mediocrely on these complex tasks, verifying the advantages of LSM in approximating complex mappings of PDEs.

\vspace{-5pt}
\paragraph{Ablations.} To verify the effectiveness of each component in LSM, we provide detailed ablations, covering both removing components (\emph{w/o}) and replacing projector (\emph{rep}) experiments. From Table \ref{tab:ablation}, we have the following observations.

In removing experiments, we can find that all components are essential to the final performance. Without the projector, model performance on both benchmarks will drop seriously, demonstrating the necessity of solving PDEs in latent space. Besides, the neural spectral block also reduces the estimation error significantly: 13.8\% (0.0253$\to$0.0218) in Elasticity-P and 13.3\% (0.0075$\to$0.0065) in Darcy. We can also find that the multiscale design can fit the Darcy benchmark well and the patchify operation is essential to the Elasticity-P benchmark, where the former always presents the multiphase flow and the latter mainly relies on the local information, showing that LSM can cover physical states in different scales and regions adaptively.

\begin{table}[t]
    \vspace{-5pt}
	\caption{Ablations on hierarchical projection network (\emph{Projection, Multiscale, Patchify}) and neural spectral block (\emph{Spectral}). We conduct two types of experiments: \emph{replacing our attention-based projector with other designs (rep)} and \emph{removing components (w/o)}. Efficiency is calculated on inputs with size $256\times 256$ and batch size as $1$. See Appendix \ref{sec:full_ablations} for full results.
 }
	\label{tab:ablation}
	\vskip 0.1in
	\centering
	\begin{small}
		\begin{sc}
			\renewcommand{\multirowsetup}{\centering}
			\setlength{\tabcolsep}{2.1pt}
			\scalebox{1}{
			\begin{tabular}{l|l|ccc|cc}
				\toprule
			\multicolumn{2}{c|}{\multirow{2}{*}{Designs}} & \scalebox{0.95}{\#Param} & \scalebox{0.95}{\#Mem} & \scalebox{0.95}{\#Time} & \multicolumn{2}{c}{\scalebox{0.95}{MSE}} \\
                \multicolumn{2}{c|}{} & \scalebox{0.8}{(MB)} & \scalebox{0.8}{(MB)} & \scalebox{0.8}{(s/iter)} & \scalebox{0.95}{Elas-P} & \scalebox{0.95}{Darcy} \\
			    \midrule
			     \multirow{3}{*}{rep} & Conv & 1.947 & 2.793 & 0.037 & 0.0236 & 0.0081 \\
                  & AvgPool & 1.836 & 1.748 & 0.028 & 0.0243 & 0.0077 \\
                  & Self-Attn & 2.002 & 7.188 & 0.064 & 0.0245 & 0.0082 \\
                  \midrule
                  \multirow{4}{*}{w/o} & Projector & 1.836 & 2.793 & 0.035 & 0.0563 & 0.0080 \\
                  & \scalebox{0.95}{Multiscale} & 0.079 & 1.757 & 0.020 & 0.0269 & 0.0123 \\
                  & Patchify & 2.002 & 1.748 & 0.062 & 0.0545 & 0.0068 \\
                  \cmidrule(lr){2-7}
                  & Spectral & 1.990 & 1.913 & 0.034 & 0.0253 & 0.0075 \\
                  \midrule
                  \multicolumn{2}{c|}{\textbf{ours}} & 2.002 & 1.914 & 0.041 & \textbf{0.0218} & \textbf{0.0065} \\
				\bottomrule
			\end{tabular}}
		\end{sc}
	\end{small}
	\vspace{-15pt}
\end{table}

\begin{figure*}[t]
\begin{center}
\centerline{\includegraphics[width=1.0\textwidth]{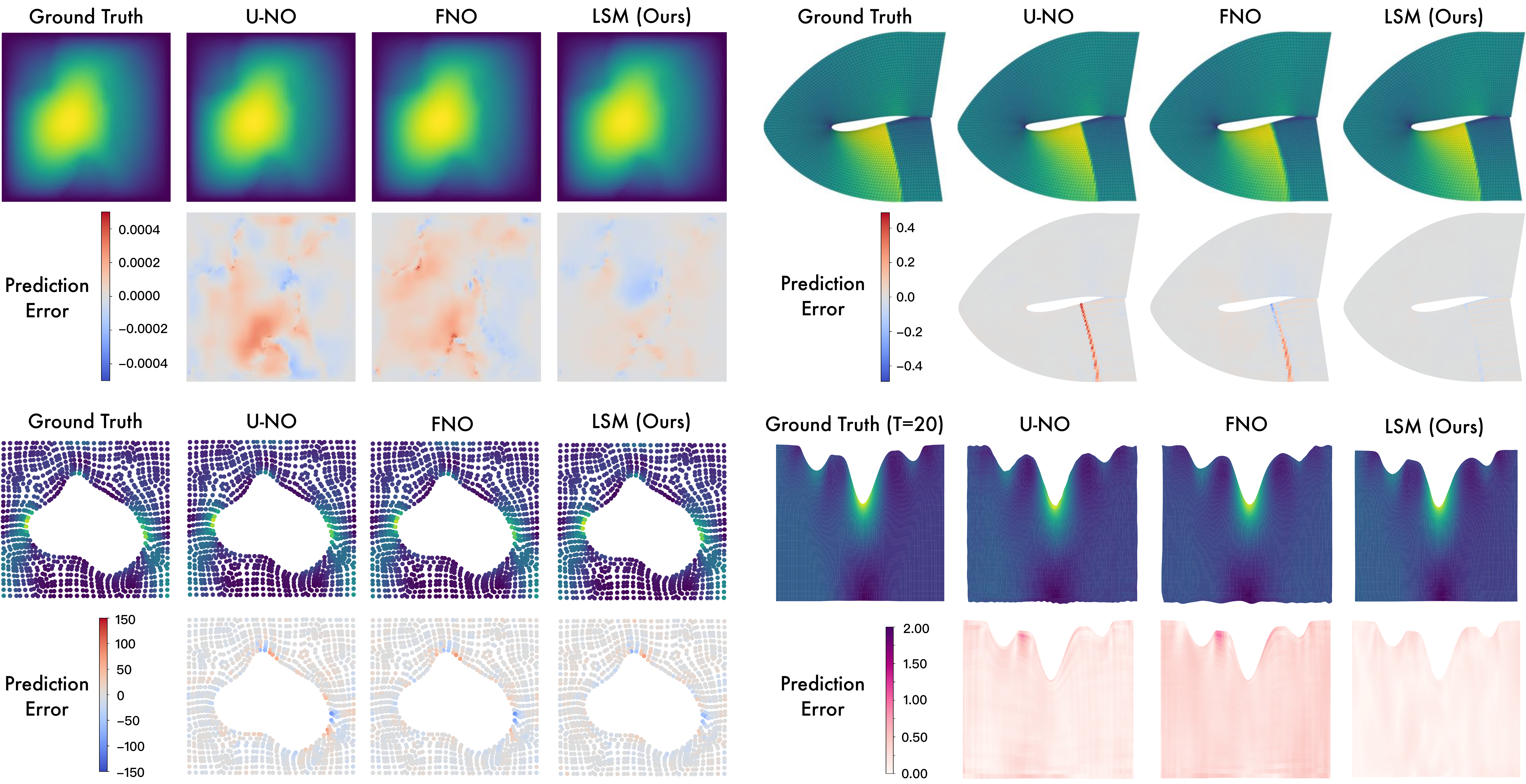}}
    \vspace{-5pt}
	\caption{Top: showcases of fluid physics on Darcy (left) and Airfoil (right); bottom: showcases of solid physics on Elasticity-P (left) and Plasticity (right). We present the last timestamp ($T=20$) for Plasticity here, which is a time-dependent task. For clearness, we also plot the prediction error, namely $\{\boldsymbol{y}(\mathbf{s})-\boldsymbol{\widehat{y}}(\mathbf{s})\}_{\mathbf{s}\in\mathcal{D}}$. See Appendix \ref{appendix:showcases} for more showcases.}
	\label{fig:case}
\end{center}
\vspace{-20pt}
\end{figure*}

In experiments of replacing our hierarchical projector, we observe that the convolution (\emph{Conv}) and canonical self-attention (\emph{Self-Attn}, \citeyear{NIPS2017_3f5ee243}) will damage both efficiency and accuracy, since they still solve PDEs in the high-dimensional coordinate space. Although average pooling (\emph{AvgPool}) can efficiently eliminate coordinate information, without latent tokens as physics prompts, it cannot capture the essential physical information and thus impairs accuracy. This verifies the efficacy of our hierarchical projection network.

\vspace{-5pt}
\paragraph{Showcases.} To present an intuitive comparison among different methods, we provide several showcases from representative benchmarks in Figure \ref{fig:case}. Generally, LSM achieves impressive performance on both solid and fluid benchmarks. Especially, for the Airfoil benchmark, LSM is the only model that precisely captures the shock wave around the airfoil, which is vital for practical design. Note that the Airfoil benchmark is to estimate the airflow velocity from the airfoil structure, where the input and output are semantically heterogeneous, demonstrating the universal approximation capacity of LSM. Besides, LSM also surpasses FNO and U-NO in estimating the inner stress of elastic materials and the future mesh deformation in plastic materials, verifying the model capability in processing complex geometrics.

\subsection{Model Analysis}
\begin{figure}[t]
\begin{center}
    \vspace{-5pt}
    \centerline{\includegraphics[width=0.5\textwidth]{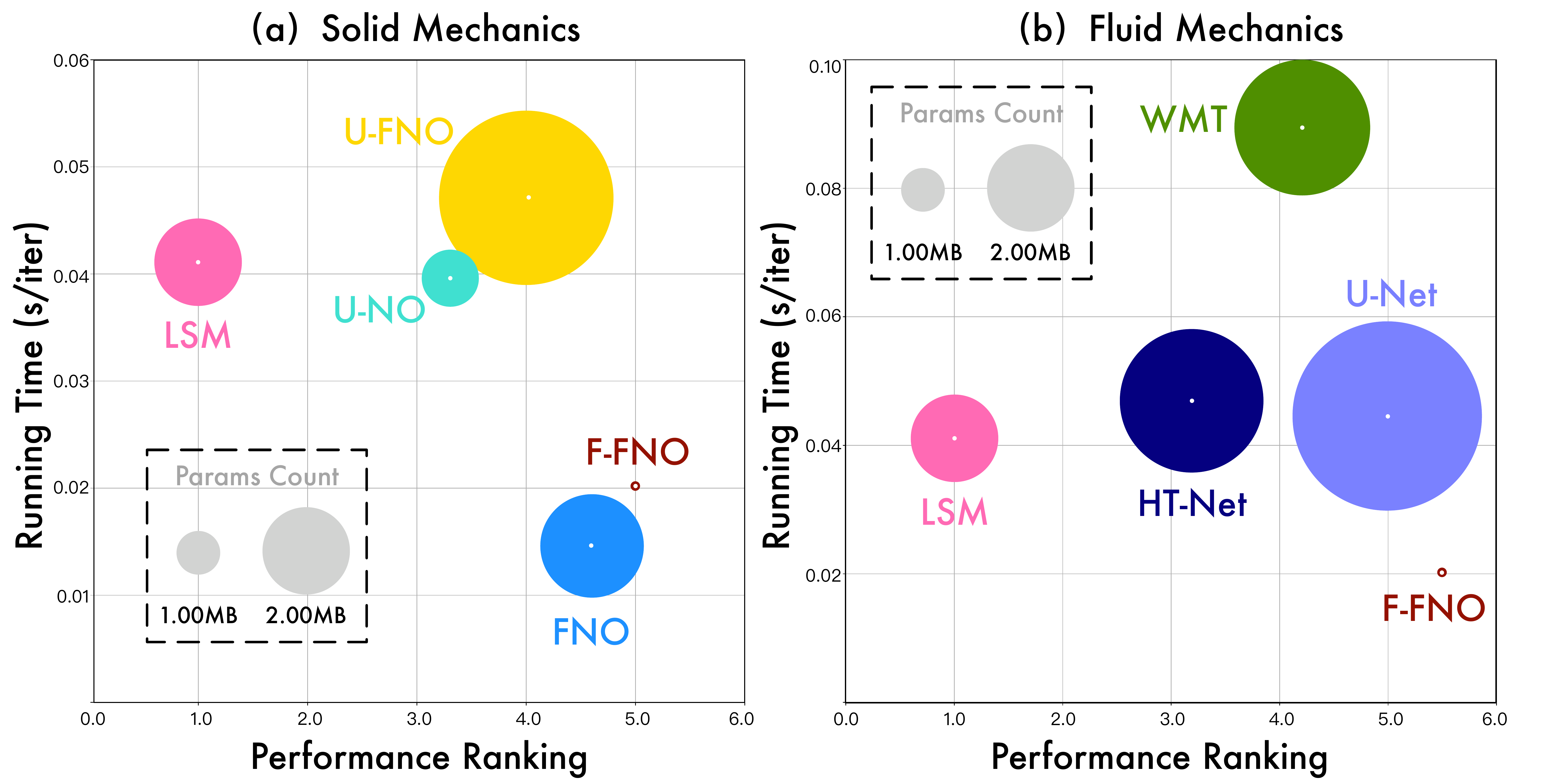}}
    \vspace{-5pt}
	\caption{Efficiency comparison for the top 5 models on the benchmarks of solid and fluid physics. Running time is evaluated on inputs with size $256\times 256$ and batch size as $1$.}
	\label{fig:efficiency}
\end{center}
\vspace{-25pt}
\end{figure}

\paragraph{Efficiency.} From Figure \ref{fig:efficiency}, we can find that LSM achieves a good trade-off between accuracy and efficiency. For solid physics, although U-NO \cite{rahman2022u} is the second-best model and slightly more efficient than LSM, LSM surpasses U-NO by a large margin, concretely 15.6\%, 12.8\% and 26.5\% relative promotion in Elasticity-P, Elasticity-G and Plasticity respectively. For fluid physics, LSM is more accurate and efficient than the previous top three baselines: HT-Net, WMT and U-Net. It is notable that F-FNO is much more lightweight than others, but its running time and accuracy are still comparable to other baselines. Thus, in comparison to the lightweight model F-FNO, LSM is still more favorable for real-world applications due to the remarkable accuracy advantage. See Table \ref{tab:all_efficiency} in Appendix for a comprehensive comparison.

\begin{figure*}[t]
\begin{center}
    \centerline{\includegraphics[width=1.0\textwidth]{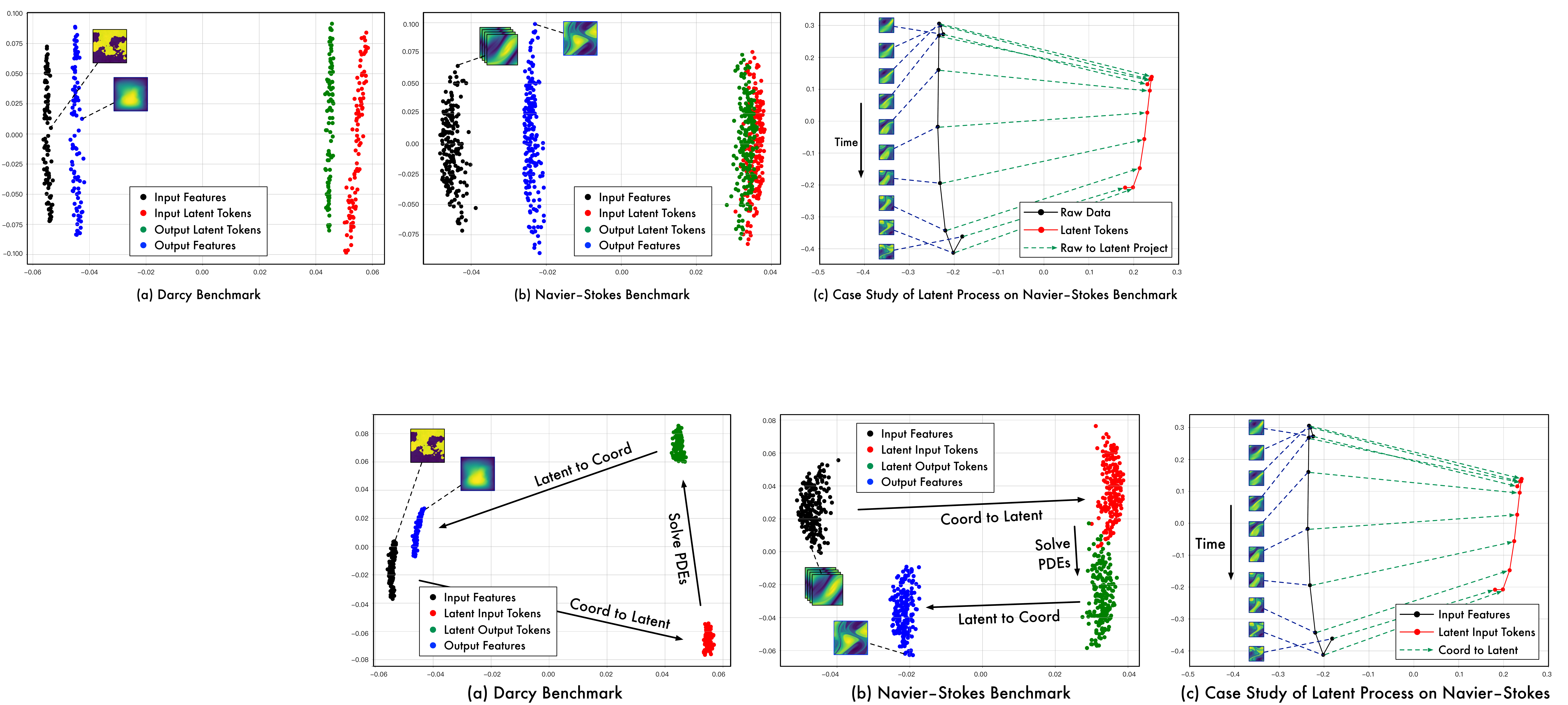}}
    \vspace{-5pt}
	\caption{Visualization of solving process. Through PCA algorithm \cite{jolliffe2016principal}, we plot the input features $\{\boldsymbol{x}(\mathbf{s})\}_{\mathbf{s}\in\mathcal{D}}$, latent input tokens $\{\mathbf{T}_{\boldsymbol{x}}\}$, latent output tokens $\{\mathbf{T}_{\boldsymbol{y}}\}$ and output features $\{\boldsymbol{{\widehat{y}}}(\mathbf{s})\}_{\mathbf{s}\in\mathcal{D}}$ into a 2D plane. All the data are from the test set.}
	\label{fig:visualization}
\end{center}
\vspace{-25pt}
\end{figure*}

\begin{table*}[t]
	\caption{Transfer the model pre-trained from full-data Pipe to limited-data Airfoil. The results are presented in the formalization of $a\to b$, where $a$ is the model performance when it is trained from scratch and $b$ is the performance finetuned from the Pipe pre-trained model. Since U-NO degenerates seriously in limited data situations, we do not take its 20\% and 40\% cases into comparison (colored in \textcolor{gray}{gray}).}
	\label{tab:transferability_all}
	\vskip 0.1in
	\centering
	\begin{small}
		\begin{sc}
			\renewcommand{\multirowsetup}{\centering}
			\setlength{\tabcolsep}{1.2pt}
			\scalebox{1}{
			\begin{tabular}{l|c|c|c|c|c}
				\toprule
			    MSE \scalebox{0.9}{($\times 10^{-2}$)} & 20\% Airfoil Data & 40\% Airfoil Data & 60\% Airfoil Data & 80\% Airfoil Data & 100\% Airfoil Data \\
			    \midrule
                U-Net \scalebox{0.7}{\citeyearpar{ronneberger2015u}} & 1.88$\to$1.93 (\textcolor{darkred}{-2.7\%}) & 1.38$\to$1.14 (\textcolor{darkblue}{+17.3\%}) & 0.96$\to$0.90 (\textcolor{darkblue}{+6.3\%}) & 0.85$\to$0.81 (\textcolor{darkblue}{+4.7\%}) & 0.79$\to$0.77 (\textcolor{darkblue}{+2.5\%}) \\
                U-NO \scalebox{0.7}{\citeyearpar{rahman2022u}} & \textcolor{gray}{6.30$\to$1.72} & \textcolor{gray}{2.39$\to$1.73} & 1.10$\to$1.00 (\textcolor{darkblue}{+9.1\%}) & 0.86$\to$0.82 (\textcolor{darkblue}{+4.7\%}) & 0.78$\to$0.82 (\textcolor{darkred}{-5.1\%}) \\
                HT-Net \scalebox{0.7}{\citeyearpar{anonymous2023htnet}} & 1.73$\to$1.43 (\textcolor{darkblue}{+17.3\%}) & 1.08$\to$0.82 (\textcolor{darkblue}{\textbf{+24.1\%}}) & 0.75$\to$0.69 (\textcolor{darkblue}{+8.0\%}) & 0.70$\to$0.65 (\textcolor{darkblue}{+7.1\%}) & 0.65$\to$0.61 (\textcolor{darkblue}{+6.2\%}) \\
                \midrule
                \textbf{LSM} & \textbf{1.66$\to$1.31} (\textcolor{darkblue}{\textbf{+21.1\%}}) & \textbf{0.91$\to$0.75} (\textcolor{darkblue}{+17.6\%}) & \textbf{0.69$\to$0.61} (\textcolor{darkblue}{\textbf{+11.6\%}}) & \textbf{0.63$\to$0.58} (\textcolor{darkblue}{\textbf{+7.9\%}}) & \textbf{0.59$\to$0.55} (\textcolor{darkblue}{\textbf{+6.8\%}}) \\
				\bottomrule
			\end{tabular}}
		\end{sc}
	\end{small}
	\vspace{-10pt}
\end{table*}

\begin{figure}[t]
\begin{center}
    \centerline{\includegraphics[width=0.5\textwidth]{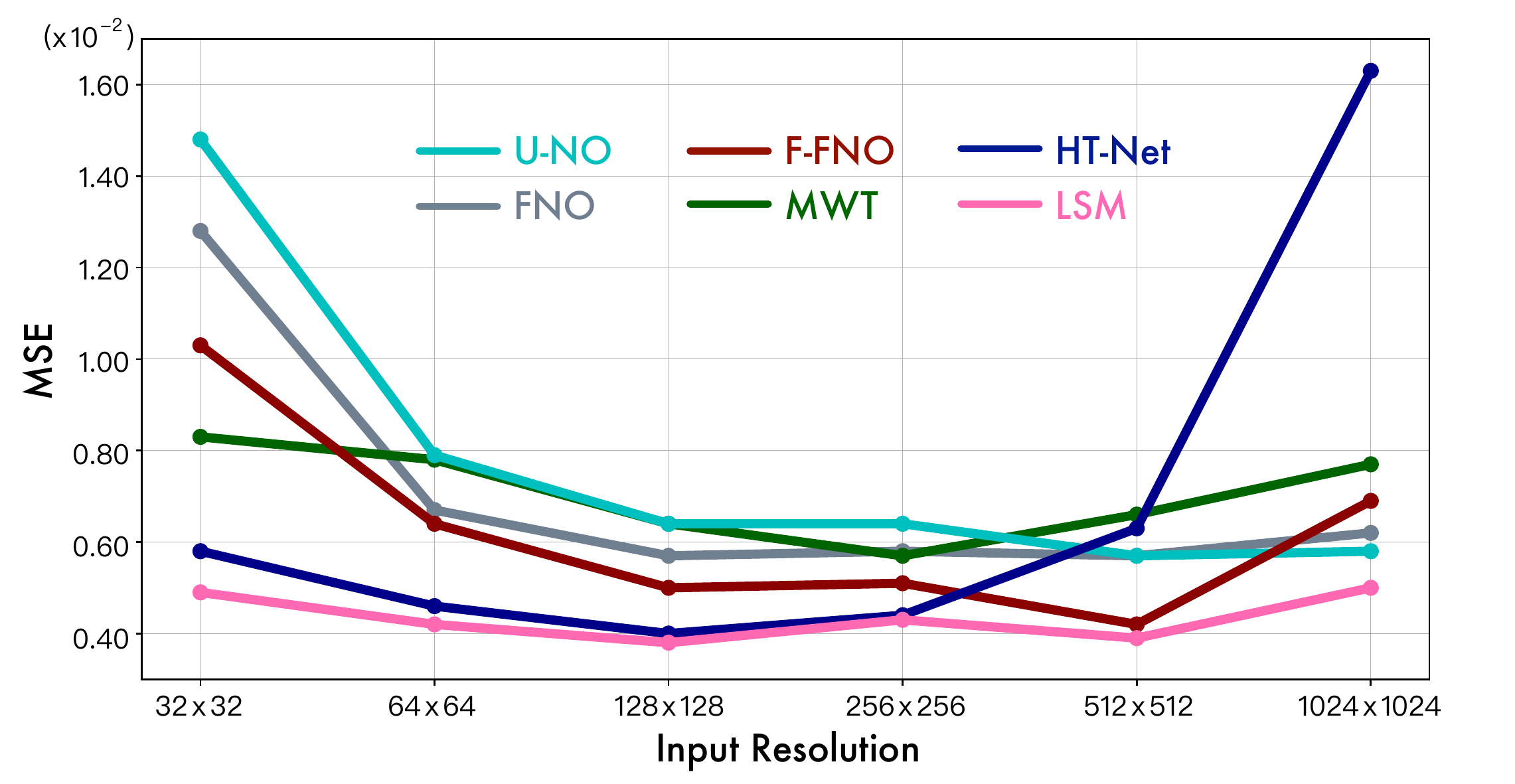}}
    \vspace{-10pt}
	\caption{Model performance of Darcy under different resolutions.}
	\label{fig:resolution}
\end{center}
\vspace{-25pt}
\end{figure}

\vspace{-5pt}
\paragraph{Solving process visualization.} We visualize the solving process of LSM in Figure \ref{fig:visualization}. From Figure \ref{fig:visualization}(a) and (b), we can easily recognize the projection and the PDE-solving process. Especially, for the Darcy benchmark, whose input and output are semantically heterogeneous, empowered by neural spectral block, LSM can present a distinct transformation in latent space to capture this complex mapping. Besides, we also provide a case for the time-dependent task from the Navier-Stokes benchmark. As shown in Figure \ref{fig:visualization}(c), by plotting the learned features over time, we can find that the latent input tokens present a similar process as the input features, demonstrating that LSM can precisely capture the latent process from high-dimensional coordinate space.

\vspace{-5pt}
\paragraph{Performance under various resolutions.} We also evaluate the model performance on the Darcy benchmark with various resolutions ranging from $32\times 32$ to $1024\times 1024$ in Figure \ref{fig:resolution}. LSM presents a stable performance w.r.t.~different inputs and consistently surpasses other baselines in all resolutions, presenting good capacity in solving high-dimensional PDEs. Besides, it is also notable that HT-Net degenerates in extremely high-dimensional setting, which is presented as a hierarchical Transformer, while FNO and its variants perform well. This indicates that there exist complex mappings between input-output pairs of high-dimensional PDEs, where even the most advanced deep models may fail without specific designs for mapping approximation.

\vspace{-5pt}
\paragraph{Transferability.} As shown in Table \ref{tab:transferability_all}, we evaluate the model transferability by finetuning the model trained on Pipe to Airfoil. We can find that LSM consistently presents the positive transfer under all limited data situations, which is meaningful for applications. Besides, it is also observed that LSM performs best in both with and without pre-training cases. Note that both two benchmarks are governed by Navier-stokes equations but with distinct boundary conditions, indicating that LSM can learn the intrinsic physical information from unwieldy high-dimensional data.

\section{Conclusions and Future Work}
In this paper, we present LSM for solving high-dimensional PDEs. Instead of directly solving PDEs in coordinate space, LSM can efficiently reduce the high-dimensional data into compact latent space by a hierarchical projection network and approximate complex mappings by neural spectral block under theoretical guarantees. Benefiting from the above designs, LSM achieves consistent state-of-the-art in both solid and fluid benchmarks and presents a good trade-off between accuracy and efficiency, making itself a promising PDE solver for real-world applications. In the future, we further explore the generalization capability of LSM among different PDEs to pursue a foundation model.

\vspace{-5pt}
\section*{Acknowledgements}
\vspace{-3pt}
This work was supported by the National Key Research and Development Plan (2021YFC3000905), National Natural Science Foundation of China (62022050 and 62021002), and Beijing Nova Program (Z201100006820041).

\bibliography{example_paper}
\bibliographystyle{icml2022}

\newpage
\appendix
\onecolumn

\section{Proofs of Theorems \ref{thm:universalappsingle}}\label{appendix:proof}
First, we would like to present a well-established theorem, whose proof can be found in the cited paper.

\begin{theorem}\label{thm:universalappsingle_general}\cite{nayak2014estimate}
Let $f:\mathbb{R}\to\mathbb{R}$ be a $2\pi$-periodic function. Its trigonometric approximation $f^{N}$ is defined as:
\begin{equation}
	\begin{split}\label{equ:singleseries}
f^{N}({x})=\sum_{k=-N}^{N}\left(\frac{1}{2\pi}\int_{-\pi}^{\pi}f(t)e^{-ik{t}}\mathrm{d}{{t}} \right)e^{i{k}{x}},
	\end{split}
\end{equation}
If $f$ satisfies the Lipschitz condition, then there is a constant $K$ that does not depend on $f$ nor $N$, such that:
\begin{equation}
	\begin{split}\label{equ:convergencespeedsingle_appendix}
|f(x)-f^{N}(x)|\leq \frac{K\ln N}{{N}},\forall x\in\mathbb{R}.
	\end{split}
\end{equation}
\end{theorem}

\begin{lemma}\label{lemma:Lipschitz}
Given $f:[0,\pi]\to\mathbb{R}$ and $g(x)=f(x)-x,\forall x\in[0,\pi]$. If $f$ satisfies the Lipschitz condition, then $g$ also satisfies the Lipschitz condition.
\end{lemma}
\begin{proof}
Suppose that $f$ satisfies the Lipschitz condition, then there is a constant $K$, such that
\begin{equation*}
\begin{split}
|f(x)-f(y)|\leq K|x-y|, \forall x,y\in [0,\pi]. 
\end{split}
\end{equation*}
Then, we have the following inequations:
\begin{equation*}
\begin{split}
|g(x)-g(y)|=|f(x)-x-\left(f(y)-y\right)|\leq |f(x)-f(y)|+|x-y|\leq (K+1)|x-y|, \forall x,y\in [0,\pi].
\end{split}
\end{equation*}
Thus, $g$ also satisfies the Lipschitz condition.
\end{proof}

\begin{lemma}\label{lemma:Lipschitz_even}
Given $f:[-\pi,\pi]\to\mathbb{R}$ and $f(x)=f(-x),\forall x\in[0,\pi]$. If $f$ satisfies the Lipschitz condition within $[0,\pi]$, then $f$ also satisfies Lipschitz condition in $[-\pi,\pi]$.
\end{lemma}
\begin{proof}
Suppose that $f$ satisfies the Lipschitz condition in $[0,\pi]$, then there is a constant $K$, such that
\begin{equation*}
\begin{split}
|f(x)-f(y)|\leq K|x-y|, \forall x,y\in [0,\pi]. 
\end{split}
\end{equation*}
$\forall x,y\in[-\pi,\pi]$, if $xy\ge0$, we obvisouly have $|f(x)-f(y)|\leq K|x-y|$. 

If $xy<0$, we have $|f(x)-f(y)|= |f(x)-f(-y)| \leq K|x+y| \leq K|x-y|$. 
\end{proof}

Next, we will prove Theorem \ref{thm:universalappsingle}, which shows the convergence property of trigonometric approximation with residual.
\begin{proof}
For simplification, we define $g(x)=f(x)-x,\forall x\in[0,\pi]$. From Lemma \ref{lemma:Lipschitz}, $g$ holds the Lipschitz condition as $f$. Then we would like to extend $g:[0,\pi]\to\mathbb{R}$ to a 2$\pi$-periodic function $g_{\text{extend}}:\mathbb{R}\to\mathbb{R}$. Firstly, we define $\widehat{g}_{\text{extend}}: [-\pi,\pi]\to\mathbb{R}$ as:
\begin{equation}
	\begin{split}\label{equ:extend_multiple}
\widehat{g}_{\text{extend}}(x)=\begin{cases}
    g(x),&\text{If $x\in[0,\pi]$}\\
    g(-x),&\text{If $x\in[-\pi,0)$},
\end{cases}
	\end{split}
\end{equation}
Further, we define the 2$\pi$-periodic function $g_{\text{extend}}:\mathbb{R}\to\mathbb{R}$ as follows:
\begin{equation}
	\begin{split}\label{equ:extend_multiple_new}
g_{\text{extend}}(x)&=\widehat{g}_{\text{extend}}\Big(\operatorname{Normalize}(x)\Big), \ \text{where} \\
\operatorname{Normalize}(x)&=\begin{cases}
            x-\operatorname{sgn}(x)\big(\lceil\frac{|x|-\pi}{2\pi}\rceil\times 2\pi\big), & \text{if $|x|>\pi$} \\
            x, & \text{otherwise},
		  \end{cases}
	\end{split}
\end{equation}
where $\operatorname{sgn}(\cdot)$ is the sign function, whose values is $1$ for positive inputs, $-1$ for negative inputs, $0$ for zero inputs.

Considering the definition of neural spectral block in Eq.~\eqref{equ:specblock}, we can find the following parameters in the neural spectral block will satisfy Eq.~\eqref{equ:convergencespeedsingle}:
\begin{equation*}
\begin{split}
\mathbf{w}_{0}&= \Big[\frac{1}{2\pi}\int_{-\pi}^{\pi}g_{\text{extend}}(t)\mathrm{d}t\Big] \\
\mathbf{w}_{\text{sin}}&= \Big[\frac{1}{\pi}\int_{-\pi}^{\pi}g_{\text{extend}}(t)\sin(t)\mathrm{d}t, \cdots, \frac{1}{\pi}\int_{-\pi}^{\pi}g_{\text{extend}}(t)\sin(\frac{N}{2} t)\mathrm{d}t\Big] \\
\mathbf{w}_{\text{cos}}&= \Big[\frac{1}{\pi}\int_{-\pi}^{\pi}g_{\text{extend}}(t)\cos(t)\mathrm{d}t, \cdots, \frac{1}{\pi}\int_{-\pi}^{\pi}g_{\text{extend}}(t)\cos(\frac{N}{2} t)\mathrm{d}t\Big]. \\
\end{split}
\end{equation*}
Then, we have the canonical trigonometric approximation of $g_{\text{extend}}$ as $g_{\text{extend}}^{N}$, which is defined as follows: 
\begin{equation*}
\begin{split}
g_{\text{extend}}^{N}(x)=\mathbf{w}_{0} + \mathbf{w}_{\text{sin}}\begin{bmatrix}
\sin(x)\\
\vdots \\
\sin(\frac{N}{2} x)
\end{bmatrix}
+\mathbf{w}_{\text{cos}}
\begin{bmatrix}
\cos(x)\\
\vdots \\
\cos(\frac{N}{2} x) \\
\end{bmatrix}=\sum_{k=-\frac{N}{2}}^{\frac{N}{2}}\left(\frac{1}{2\pi}\int_{-\pi}^{\pi}g_{\text{extend}}(t)e^{-ik{t}}\mathrm{d}{{t}} \right)e^{i{k}{x}}, \forall x\in\mathbb{R}.
\end{split}
\end{equation*}

If $f$ satisfies the Lipschitz condition, from Lemma \ref{lemma:Lipschitz} and Lemma \ref{lemma:Lipschitz_even}, we have that $\widehat{g}_{\text{extend}}$ satisfies the Lipschitz condition. Since $\widehat{g}_{\text{extend}}:[-\pi,\pi]\to\mathbb{R}$ satisfies $\widehat{g}_{\text{extend}}(x)=\widehat{g}_{\text{extend}}(-x)$, then $\forall x,y\in\mathbb{R}$, there is a constant $K^\prime$, such that:
\begin{equation}\label{equ:Lipschitz_extend}
\begin{split}
|g_{\text{extend}}(x)-g_{\text{extend}}(y)|& = \Big|\widehat{g}_{\text{extend}}\left(|\operatorname{Normalize}(x)|\right)-\widehat{g}_{\text{extend}}\left(|\operatorname{Normalize}(y)|\right)\Big| \\ 
&= \Big|g\left(|\operatorname{Normalize}(x)|\right)-g\left(|\operatorname{Normalize}(y)|\right)\Big| \\ 
&\leq K^\prime\Big||\operatorname{Normalize}(x)|-|\operatorname{Normalize}(y)|\Big| \\
& \leq K^\prime|x-y|.\quad \textcolor{gray}{\text{(Similar discussion as Lemma \ref{lemma:Lipschitz_even})}}
\end{split}
\end{equation}
For the last inequation of Eq.~\eqref{equ:Lipschitz_extend}, if $|x-y|\ge\pi$, the inequation obviously holds. If $|x-y|<\pi$ and $x,y\in[n\pi, (n+1)\pi], n\in\mathbb{Z}$, then we have $\Big||\operatorname{Normalize}(x)|-|\operatorname{Normalize}(y)|\Big|=|x-y|$. As for $|x-y|<\pi$ and $x\leq 2n\pi\leq y, n\in\mathbb{Z}$ (suppose $x\leq y$ without loss of generality), we have \begin{equation*}
\begin{split}
\Big||\operatorname{Normalize}(x)|-|\operatorname{Normalize}(y)|\Big|=\Big|(2n\pi-x)-(y-2n\pi)\Big|\leq\Big|(2n\pi-x)+(y-2n\pi)\Big|=|x-y|.
\end{split}
\end{equation*}
As for $|x-y|<\pi$ and $x\leq (2n+1)\pi\leq y, n\in\mathbb{Z}$, we have
\begin{equation*}
\begin{split}
\Big||\operatorname{Normalize}(x)|-|\operatorname{Normalize}(y)|\Big|& =\Big|\big(\pi-((2n+1)\pi-x)\big)-\big(\pi-(y-(2n+1)\pi)\big)\Big| \\
& =\Big|(y-(2n+1)\pi)-((2n+1)\pi-x)\Big| \\
& \leq\Big|(y-(2n+1)\pi)+((2n+1)\pi-x)\Big| \\
& = |x-y|.
\end{split}
\end{equation*}
Thus, $g_{\text{extend}}$ also satisfies the Lipschitz condition.

Thus, from Theorem \ref{thm:universalappsingle_general}, we have that $g_{\text{extend}}^N$ converges to $g_{\text{extend}}$ with the speed as follows:
\begin{equation}
	\begin{split}\label{equ:convergencespeedsingle_g}
|g_{\text{extend}}(x)-g_{\text{extend}}^{N}(x)|\leq \frac{K\ln \frac{N}{2}}{\frac{N}{2}},\forall x\in\mathbb{R},
	\end{split}
\end{equation}
where $K$ is a constant. From the definition of Eq.~\eqref{equ:specblock}, $\forall x\in[0,\pi]$, we have $f^{N}(x)=x+g_{\text{extend}}^{N}(x)$, then
\begin{equation}
	\begin{split}\label{equ:convergencespeedsingle_final}
|f(x)-f^{N}(x)|&=\bigg|\big(g_{\text{extend}}(x)+x\big)-\big(g_{\text{extend}}^{N}(x)+x\big)\bigg| \\
&=\bigg|g_{\text{extend}}(x)-g_{\text{extend}}^{N}(x)\bigg| \leq \frac{K\ln \frac{N}{2}}{{\frac{N}{2}}}\leq \frac{(2K)\ln N}{N}, \forall x\in[0,\pi].
	\end{split}
\end{equation}
\end{proof}

\section{Details for Benchmarks}\label{appendix:dataset}
We have summarized benchmark configurations in Table \ref{tab:dataset_detail}. Here are the generation details categorized by governing PDEs.

\begin{table*}[t]
	\caption{Details for benchmarks. All the settings follow FNO \cite{li2021fourier} and geo-FNO \cite{Li2022FourierNO}. The input-output resolutions are presented in the shape of (temporal, spatial, variate). ``/'' means without this dimension.}
	\label{tab:dataset_detail}
	\vspace{-5pt}
	\vskip 0.15in
	\centering
 \resizebox{1.02\columnwidth}{!}
 {
 \begin{threeparttable}
	\begin{small}
		\begin{sc}
			\renewcommand{\multirowsetup}{\centering}
			\setlength{\tabcolsep}{2pt}
			\begin{tabular}{ll|c|c|c|c|c|c|c}
				\toprule
                    \multicolumn{2}{c}{\multirow{3}{*}{Descriptions}} & \multicolumn{3}{c}{Solid Physics} & \multicolumn{4}{c}{Fluid Physics} \\
                    \cmidrule(lr){3-5}\cmidrule(lr){6-9}
				 & & Elasticity-P & Elasticity-G & Plasticity & Navier–Stokes  & Airfoil & Pipe & Darcy \\
                \midrule
		\multirow{6}{*}{\rotatebox{90}{Physics}} & PDEs & \multicolumn{3}{c|}{PDEs of solid material} & \multicolumn{3}{c|}{Navier-Stokes Equation} & Darcy’s law \\
                \cmidrule(lr){2-9}
			 & Task & \multicolumn{2}{c|}{Estimate stress} & Model deformation & Predict future & \multicolumn{2}{c|}{Estimate velocity} & Estimate Pressure\\
                \cmidrule(lr){2-9}
		   & Input & \multicolumn{2}{c|}{Material structure} & Boundary condition & Past velocity & \multicolumn{2}{c|}{Structure} & Porous medium \\
                \cmidrule(lr){2-9}
			& Output & \multicolumn{2}{c|}{Inner stress} & Mesh displacement & Future velocity & \multicolumn{2}{c|}{Fluid velocity} & Fluid Pressure \\
                \midrule
			\multirow{6}{*}{\rotatebox{90}{Data}} & Train set size & 1000 & 1000 & 900 & 1000 & 1000 & 1000 & 1000 \\
                \cmidrule(lr){2-9}
			 & Test set size & 200 & 200 & 80 & 200 & 100 & 200 & 200 \\
                \cmidrule(lr){2-9}
			 & Input Tensor & $(/, 972, 2)$ & $(/, 41\times41, 1)$ & $(/, 101\times31, 2)$ & $(10, 64\times64, 1)$ & $(/, 200\times50, 2)$ & $(/, 129\times129, 2)$ & $(/, 85\times85, 1)$ \\
                \cmidrule(lr){2-9}
			 & Output Tensor & $(/, 972, 1)$ & $(/, 41\times41, 1)$ & $(20, 101\times31, 4)$ & $(10, 64\times64, 1)$ & $(/, 200\times50, 1)$ & $(/, 129\times129, 1)$ & $(/, 85\times85, 1)$ \\
				\bottomrule
			\end{tabular}
		\end{sc}
	\end{small}
 \end{threeparttable}
    }
\end{table*}

\subsection{Solid Material} The governing equation of solid material is:
\begin{equation}\label{equ:solid_pde}
    \rho^{s}\frac{\partial^2 \boldsymbol{u}}{\partial t^2} + \nabla \cdot  \boldsymbol{\sigma}=0,
\end{equation}
where $\rho^s\in\mathbb{R}$ means the solid density, $\nabla$ denotes the nabla operator. $\boldsymbol{u}$ is a function that represents the displacement vector of material over time $t$. $\boldsymbol{\sigma}$ denotes the stress tensor. Elasticity-P, Elasticity-G and Plasticity \cite{Li2022FourierNO} share the same governing equation as shown in Eq.~\eqref{equ:solid_pde}.

\paragraph{Elasticity-P and Elasticity-G.} These benchmarks are to estimate the inner stress of an incompressible material with an arbitrary void at the center of the material. Besides, an external tension is applied to the material. The input is the structure of the material, and the output is inner stress. Elasticity-P and Elasticity-G differ in the way modeling the geometric of material: Elasticity-P uses a point cloud with 972 points, while Elasticity-G presents the data in a regular grid with the size of $41\times41$, which is interpolated from Elasticity-P.

\paragraph{Plasticity.} This benchmark focuses on the plastic forging problem, where a plastic material is impacted from above by an arbitrary-shaped die. The input is the shape of the die, which is recorded in structured mesh. And the output is the deformation of each mesh point in the future $20$ time steps. The resolution of the structured mesh is $101\times31$.

\subsection{Navier-Stokes Equation}
The differential form of fluid dynamics equations are:
\begin{align}
    \label{equ:mass}\frac{\partial\rho}{\partial t} + \nabla\cdot(\rho\boldsymbol{U})&=0\\
    \label{equ:momentum}\frac{\partial\boldsymbol{U}}{\partial t} + \boldsymbol{U}\cdot\nabla\boldsymbol{U}&=\boldsymbol{f}+\frac{1}{\rho}\nabla\cdot(\boldsymbol{T}_{ij}\boldsymbol{e}_i\boldsymbol{e}_j)\\
    \label{equ:energy}\frac{\partial(e+\frac{1}{2}\boldsymbol{U}^2)}{\partial t} + \boldsymbol{U}\cdot\nabla(e+\frac{1}{2}\boldsymbol{U}^2)&=\boldsymbol{f}\cdot\boldsymbol{U}+\frac{1}{\rho}\nabla\cdot(\boldsymbol{U}\cdot \boldsymbol{T}_{ij}\boldsymbol{e}_i\boldsymbol{e}_j)+\frac{\lambda}{\rho}\Delta T,
\end{align}
where Eq.~\eqref{equ:mass}, Eq.~\eqref{equ:momentum} and Eq.~\eqref{equ:energy} describe the mass, momentum and energy conservation respectively. Here $\rho$ is the density, $\boldsymbol{U}$ is the velocity vector, $\boldsymbol{f}$ is the external force, $e$ is the internal energy. And $\boldsymbol{T}$ is the stress tensor in the fluid, $\boldsymbol{e}$ is the basis vector and $\boldsymbol{T}_{ij}\boldsymbol{e}_{i}\boldsymbol{e}_j$ follows the Einstein summation convention. All above variates are related to both space and time. $\frac{\lambda}{\rho}\Delta T$ is for heat conduction. For a Newtonian fluid, the stress tensor $\boldsymbol{T}$ is related to the pressure $p$, viscosity coefficient $\nu$ and velocity vector $\boldsymbol{U}$. Thus, for the Newtonian fluid, Eq.~\eqref{equ:momentum} can be rewritten as:
\begin{equation}
    \begin{split}\label{equ:NS}
      \frac{\partial\boldsymbol{U}}{\partial t} + \boldsymbol{U}\cdot\nabla\boldsymbol{U} &= \boldsymbol{f} - \frac{1}{\rho}\nabla p + \nu\nabla^2\boldsymbol{U}.
    \end{split}
\end{equation}
Besides, Eq.~\eqref{equ:energy} can also be deduced in a similar way, but the result is too complex to be presented in this paper. See \cite{mclean2012continuum} for more details. The dynamics equations for Newtonian fluid are well-known as Navier-Stokes equations. Next, we will detail the underlying PDEs for our fluid benchmarks.

\paragraph{Navier-Stokes.} We take the Navier-Stokes dataset from \cite{li2021fourier}. This dataset simulates incompressible and viscous flow on the unit torus, where the density of fluid is unchangeable ($\rho$ in Eq.~\eqref{equ:mass}). In this situation, the energy conservation presented in Eq.~\eqref{equ:energy} is independent of mass and momentum conservation. Hence, the fluid dynamics can be deduced with Eq.~\eqref{equ:mass} and Eq.~\eqref{equ:NS}:
\begin{equation}
\begin{split}
    \nabla \cdot \boldsymbol{U} &= 0\\
    \frac{\partial w}{\partial t} + \boldsymbol{U} \cdot \nabla w &= \nu \nabla^2 w + f\\
    w|_{t=0}&=w_0,
\end{split}
\end{equation}
where $\boldsymbol{U}=(u,v)$ is a velocity vector in 2D field, $w=|\nabla \times \boldsymbol{U}|=\frac{\partial u}{\partial y}-\frac{\partial v}{\partial x}$ is the vorticity, $w_0\in \mathbb{R}$ is the initial vorticity  at $t=0$. In this dataset, viscosity $\nu$ is set as $10^{-5}$ and the resolution of the 2D field is $64\times64$. Each generated sample contains 20 successive frames and the task is to predict the future 10 frames based on the past 10 frames.

\paragraph{Pipe.} This dataset \cite{Li2022FourierNO} focuses on the incompressible flow through a pipe. The governing equations are similarly deduced with Eq.~\eqref{equ:mass} and Eq.~\eqref{equ:NS}:
\begin{equation}
    \begin{split}
      \nabla \cdot \boldsymbol{U}&=0\\
      \frac{\partial\boldsymbol{U}}{\partial t} + \boldsymbol{U}\cdot\nabla\boldsymbol{U} &= \boldsymbol{f} - \frac{1}{\rho}\nabla p + \nu\nabla^2\boldsymbol{U}.
    \end{split}
\end{equation}
The dataset is generated in the geometric of structured mesh with the resolution of $129\times129$. For experiments, we adopt the mesh structure as the input data, and the output is the horizonal fluid velocity within the pipe.

\paragraph{Airfoil.} The airfoil dataset \cite{Li2022FourierNO} is about the transonic flow over an airfoil. Since the viscosity of air is quite small, the viscous term $\nu\nabla^2 \boldsymbol{U}$ can be ignored in the Navier-Stokes equation. Thus, the governing equations for this situation can be presented as follows:
\begin{equation}
    \begin{split}
        \frac{\partial \rho^f}{\partial t} + \nabla\cdot(\rho^f \boldsymbol{U})&=0\\
        \frac{\partial\rho^f\boldsymbol{U}}{\partial t}+\nabla\cdot(\rho^f\boldsymbol{U}\boldsymbol{U}+p\mathbb{I})&=0\\
        \frac{\partial E}{\partial t}+\nabla\cdot((E+p)\boldsymbol{U})&=0,
    \end{split}
\end{equation}
where $\rho^f$ denotes the fluid density, and $E$ represents the total energy. The data is generated in the geometric of structured mesh with resolution of $200\times50$. The locations of these mesh points are adopted as inputs. And the Mach number of each mesh point is the output.

\subsection{Darcy Flow}
\paragraph{Darcy.} The Darcy's law describes the flow of fluid through a porous medium, for example, water goes through sand. We use the Darcy dataset proposed in \cite{li2021fourier}, where 2-D Darcy flow equations in a unit box are formulized as:
\begin{equation}
\begin{split}
    -\nabla \cdot (a\nabla u)&=f\\
    u|_{x\in \partial (0,1)^2}&=0,
\end{split}
\end{equation}
where $a\in\mathbb{R}^+$ is the diffusion coefficient. $f$ means the externel force, which is fixed as $1$ in this dataset. This dataset takes $a$ as input, and the output is the solution $u$. The samples in this dataset are in the regular grid with resolution as $85\times85$.

\section{More Showcases} \label{appendix:showcases}
\begin{figure*}[t]
\begin{center}
    \centerline{\includegraphics[width=1.0\textwidth]{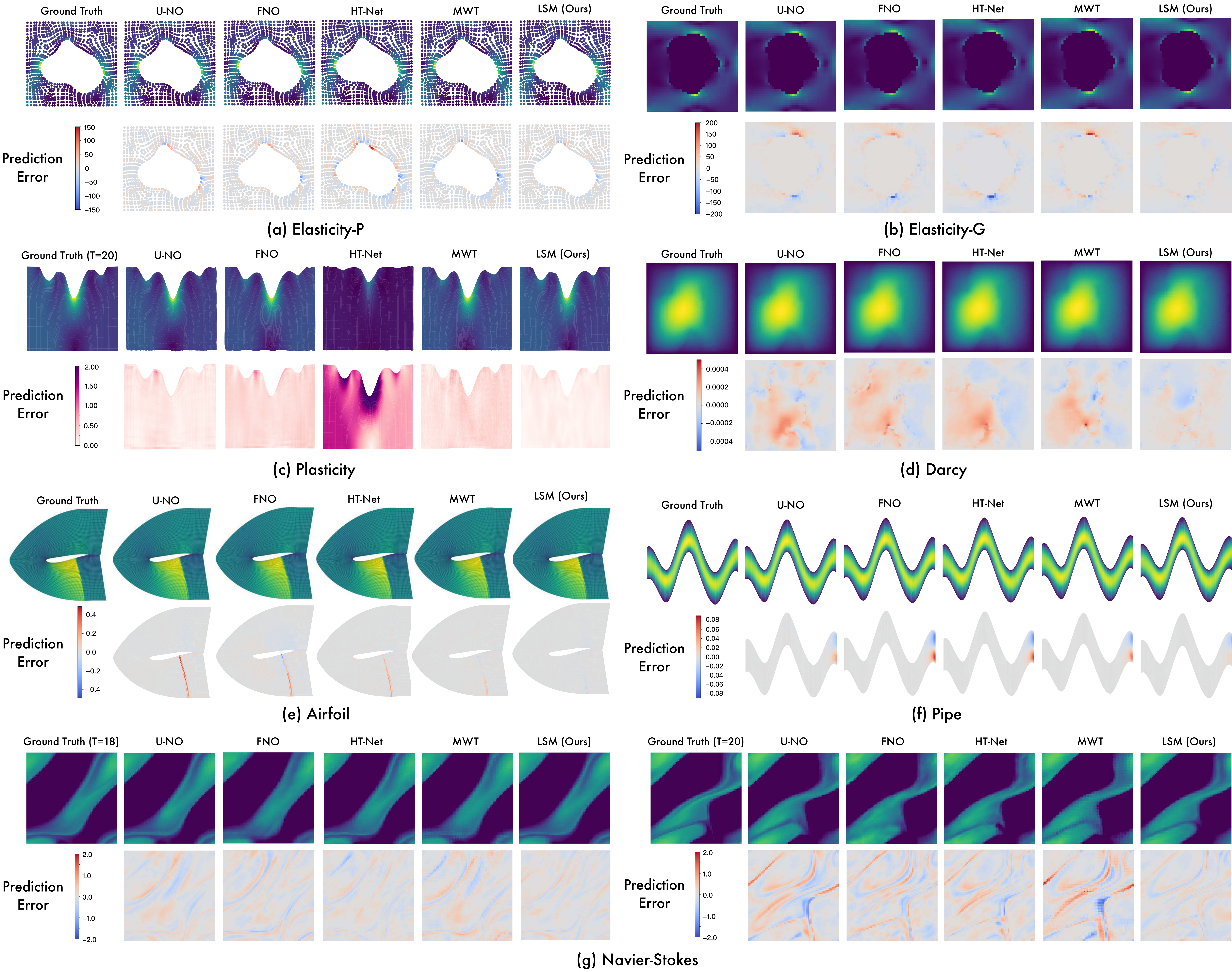}}
    \vspace{-5pt}
	\caption{Showcases on all seven benchmarks. Especially for the sub-figure (c) Plasticity, we plot the last timestamp of output ($T=20$). As for the sub-figure (g) Navier-Stokes, we plot the frames at $T=18$ and $T=20$ to present the model performance over time.}
	\label{fig:case_all}
\end{center}
\vspace{-20pt}
\end{figure*}

As a complement to Figure \ref{fig:case}, we present showcases for all benchmarks in Figure \ref{fig:case_all} and also plot the coordinate-wise prediction error for comparison. As demonstrated in above showcases, LSM achieves a remarkable prediction performance in extensive tasks. By investigating each case, we can obtain the following observations:
\begin{itemize}
    \item Performance on the boundary. From Figure \ref{fig:case_all}(a)(b)(c)(f), we can find that LSM significantly surpasses other baselines on the boundary of different geometrics, demonstrating the model capability in learning physical constraints.
    \item Performance in time-dependent tasks.
    As shown in Figure \ref{fig:case_all}(g), LSM can precisely predict the future velocity for the fluid in the Navier-Stokes benchmark. Especially, the performances of other methods drop seriously from $T=18$ to $T=20$, while LSM can simulate the fluid accurately even in the long-term future.
\end{itemize}

\section{Full Ablations}\label{sec:full_ablations}
As a complement to Table \ref{tab:ablation} of main text, we provide the comprehensive ablation results for all seven benchmarks here. From Table \ref{tab:ablation_full_res}, we can observe that all the components in LSM are effective to the final performance. Besides, we also present detailed ablations on the neural spectral block in Table \ref{tab:ablation_neural_spectral}. Here are the analyses.

\paragraph{Replace neural spectral block with other global operators.} To verify advantages in learning multiple basis operators, we also conduct experiments on replacing the neural spectral block with multilayer perceptrons (MLP) and FNO \cite{li2021fourier}, where the latter ones are global operators without basis operator decomposition design. As shown in Table \ref{tab:ablation_neural_spectral}, compared to learning basis operators, it is harder to learn a global operator, whose performance is close to removing the neural spectral block directly. It is also notable that replacing neural spectral block with FNO means applying FFT in the latent space, which is unreasonable since the latent tokens are independent. Thus, directly replacing neural spectral block with FNO damages performance seriously (Table \ref{tab:ablation_neural_spectral}), sometimes even worse than the case without neural spectral block.

\paragraph{Replace basis operators in neural spectral block.} Note that the classical spectral method is a general framework, which is to decompose the complex solution into several orthogonal basis functions. Thus, replacing the trigonometric approximation in LSM with other basis is also implementable. However, other basis may not achieve the nice approximation and optimization properties as the trigonometric basis functions. For example, as shown in Table \ref{tab:ablation_neural_spectral}, directly replacing trigonometric basis with polynomial basis will decrease the model performance. Thus, we would like to leave the exploration of other basis operators as the future work, including the corresponding model design and theoretical derivation.

\begin{table}[t]
    \vspace{-5pt}
	\caption{Full ablation results on hierarchical projection network (\emph{Projection, Multiscale, Patchify}) and neural spectral block (\emph{Spectral}). We conduct two types of experiments: \emph{replacing our attention-based projector with other designs (rep)} and \emph{removing components (w/o)}. Efficiency is calculated on inputs with size $256\times 256$ and batch size as $1$. ``/'' indicates the out-of-memory situation.
 }
	\label{tab:ablation_full_res}
	\vskip 0.1in
	\centering
	\begin{small}
		\begin{sc}
			\renewcommand{\multirowsetup}{\centering}
			\setlength{\tabcolsep}{1.8pt}
			\scalebox{1}{
			\begin{tabular}{l|l|ccc|ccccccc}
				\toprule
			\multicolumn{2}{c|}{\multirow{2}{*}{Designs}} & \scalebox{0.95}{\#Param} & \scalebox{0.95}{\#Mem} & \scalebox{0.95}{\#Time} & \multicolumn{3}{c}{\scalebox{0.95}{Solid Physics}} & \multicolumn{4}{c}{\scalebox{0.95}{Fluid Physics}} \\
            \cmidrule(lr){6-8}\cmidrule(lr){9-12}
                \multicolumn{2}{c|}{} & \scalebox{0.8}{(MB)} & \scalebox{0.8}{(MB)} & \scalebox{0.8}{(s/iter)} & \scalebox{0.95}{Elasticity-P} & \scalebox{0.95}{Elasticity-G} & \scalebox{0.95}{Plasticity} & \scalebox{0.95}{Navier-Stokes} & \scalebox{0.95}{Darcy} & \scalebox{0.95}{Airfoil} & \scalebox{0.95}{Pipe} \\
			    \midrule
			     \multirow{3}{*}{rep} & Conv & 1.947 & 2.793 & 0.037 & 0.0236 & 0.00429 & 0.0029 & 0.1571 & 0.0081 & 0.0077 & 0.0052 \\
                  & AvgPool & 1.836 & 1.748 & 0.028 & 0.0243 & 0.0413 & 0.0031 & 0.1564 & 0.0077 & 0.0072 & 0.0056 \\
                  & Self-Attn & 2.002 & 7.188 & 0.064 & 0.0245 & 0.0424 & / & 0.1567 & 0.0082 & 0.0062 & 0.0056 \\
                  \midrule
                  \multirow{4}{*}{w/o} & Projector & 1.836 & 2.793 & 0.035 & 0.0563 & 0.0419 & / & 0.1609 & 0.0080 & 0.0085 & 0.0059 \\
                  & \scalebox{0.95}{Multiscale} & 0.079 & 1.757 & 0.020 & 0.0269 & 0.0479 & 0.0044 & 0.1667 & 0.0123 & 0.0097 & 0.0091 \\
                  & Patchify & 2.002 & 1.748 & 0.062 & 0.0545 & 0.0414 & 0.0040 & 0.1576 & 0.0068 & 0.0062 & 0.0055 \\
                  \cmidrule(lr){2-12}
                  & Spectral & 1.990 & 1.913 & 0.034 & 0.0253 & 0.0421 & 0.0034 & 0.1618 & 0.0075  & 0.0107 & 0.0053 \\
                  \midrule
                  \multicolumn{2}{c|}{\textbf{ours}} & 2.002 & 1.914 & 0.041 & \textbf{0.0218} & \textbf{0.0408} & \textbf{0.0025} & \textbf{0.1535} & \textbf{0.0065} & \textbf{0.0059} & \textbf{0.0050} \\
				\bottomrule
			\end{tabular}}
		\end{sc}
	\end{small}
\end{table}

\begin{table}[t]
    \vspace{-5pt}
	\caption{Detailed ablations on neural spectral block. MSE is recorded.
 }
	\label{tab:ablation_neural_spectral}
	\vskip 0.1in
	\centering
	\begin{small}
		\begin{sc}
			\renewcommand{\multirowsetup}{\centering}
			\setlength{\tabcolsep}{3pt}
			\scalebox{1}{
			\begin{tabular}{l|l|ccc}
				\toprule
			Type & Model & Elasticity-P & Darcy \\
                \midrule
                \multirow{3}{*}{Replace neural spectral block} & LSM w/o neural spectral block & 0.0253 & 0.0075 \\
                & LSM but replace neural spectral block with MLP & 0.0249 & 0.0075 \\
                & LSM but replace neural spectral block with FNO & 0.0356 & 0.0073 \\
                \midrule
                {Replace basis operators} & LSM with polynomial basis operators & 0.0261 & 0.0073 \\
                \midrule
                Final version & LSM & \textbf{0.0218} & \textbf{0.0065} \\
				\bottomrule
			\end{tabular}}
		\end{sc}
	\end{small}
	\vspace{-5pt}
\end{table}

\section{Performance Under Various Resolutions}
As shown in Table \ref{tab:resolution} and \ref{tab:resolution_ns}, we also evaluate the model performance on the newly-generated Darcy and Navier-Stokes datasets with various resolutions, where we can obtain the following observations:

\begin{itemize}
    \item For the Darcy benchmark, U-Net \citeyearpar{ronneberger2015u} and HT-Net \cite{anonymous2023htnet} that are proposed based on advanced deep models U-Net and Transformer \cite{NIPS2017_3f5ee243}, degenerate a lot on the inputs with large resolutions, e.g. $1024\times 1024$, indicating that there exist complex mappings between input-output pairs of high-dimensional PDEs. In contrast, LSM presents a stable performance w.r.t.~different inputs and consistently surpasses other baselines in all resolutions, presenting good capacity in solving high-dimensional PDEs.
    \item As for the Navier-Stokes benchmark, whose task is to predict the future 10 frames based the past 10 frames, we can find that in comparison with other baselines, LSM presents more significant advantage in higher input resolutions.
\end{itemize}

\begin{table}[h]
    \vspace{-10pt}
    \caption{Model performance comparison on Darcy under different resolutions.}
	\label{tab:resolution}
	\vskip 0.1in
	\centering
	\begin{small}
		\begin{sc}
			\renewcommand{\multirowsetup}{\centering}
			\setlength{\tabcolsep}{2.5pt}
			\scalebox{1}{
			\begin{tabular}{c|cccccc|c}
				\toprule
			    Resolution & U-Net \citeyearpar{ronneberger2015u} & FNO \citeyearpar{li2021fourier} & MWT \citeyearpar{Gupta2021MultiwaveletbasedOL} & U-NO \citeyearpar{rahman2022u} & F-FNO \citeyearpar{anonymous2023factorized} & HT-Net \citeyearpar{anonymous2023htnet} & \textbf{LSM (ours)}\\
			    \midrule
                $32\times 32$ & 0.0059 & 0.0128 & 0.0083 & 0.0148 & 0.0103 & 0.0058 & \textbf{0.0049}\\
                $64\times 64$ & 0.0052 & 0.0067 & 0.0078 & 0.0079 & 0.0064 & 0.0046 & \textbf{0.0042}\\
                $128\times 128$ & 0.0054 & 0.0057 & 0.0064 & 0.0064 & 0.0050 & 0.0040 & \textbf{0.0038}\\
                $256\times 256$ & 0.0251 & 0.0058 & 0.0057 & 0.0064 & 0.0051 & 0.0044 & \textbf{0.0043} \\
                $512\times 512$ & 0.0496 & 0.0057 & 0.0066 & 0.0057 & 0.0042 & 0.0063 & \textbf{0.0039} \\
                $1024\times 1024$ & 0.0754 & 0.0062 & 0.0077 & 0.0058 & 0.0069 & 0.0163 & \textbf{0.0050} \\
				\bottomrule
			\end{tabular}}
		\end{sc}
	\end{small}
	\vspace{-10pt}
\end{table}

\begin{table}[h]
    \caption{Model performance comparison on the Navier-Stokes benchmark under different resolutions. ``/'' indicates the poor performance.}
	\label{tab:resolution_ns}
	\vskip 0.1in
	\centering
	\begin{small}
		\begin{sc}
			\renewcommand{\multirowsetup}{\centering}
			\setlength{\tabcolsep}{2.5pt}
			\scalebox{1}{
			\begin{tabular}{c|cccccc|c}
				\toprule
			    Resolution & U-Net \citeyearpar{ronneberger2015u} & FNO \citeyearpar{li2021fourier} & MWT \citeyearpar{Gupta2021MultiwaveletbasedOL} & U-NO \citeyearpar{rahman2022u} & F-FNO \citeyearpar{anonymous2023factorized} & HT-Net \citeyearpar{anonymous2023htnet} & \textbf{LSM (ours)}\\
			    \midrule
                $64\times 64$ & 0.1982 & 0.1556 & 0.1541 & 0.1713 & 0.2322 & 0.1847 & \textbf{0.1535}\\
                $128\times 128$ & / & 0.1028 & 0.1099 & 0.1068 & 0.1506 & 0.1088 & \textbf{0.0961}\\
				\bottomrule
			\end{tabular}}
		\end{sc}
	\end{small}
\end{table}

\section{Additional Experiments on Burger's Equation}
As a fundamental partial differential equation for convection-diffusion processes occurring in various areas of applied mathematics, Burger's equation is widely used in modeling fluid mechanics, nonlinear acoustics and gas dynamics. Following the experiment settings in FNO \cite{li2021fourier}, we also test LSM in solving 1D Burger's equation. Especially, to fit the 1D input data, we need to implement the following changes to LSM and other baselines:
\begin{itemize}
    \item By conducting the up-down sampling and patchify in the 1D space, LSM can handle the 1D inputs.
    \item As for the F-FNO, we replace its factorized 2D FFT with 1D FFT.
    \item For the U-NO, we replace both 2D up-down sampling and 2D FFT with 1D versions.
\end{itemize}
From Table \ref{tab:burger}, we can find that LSM still performs well in this equation under various resolutions, verifying the model capacity in solving high-dimensional PDEs.

\begin{table}[h]
    \vspace{-10pt}
    \caption{Model performance comparison on 1D Burger's equation.}
	\label{tab:burger}
	\vskip 0.1in
	\centering
	\begin{small}
		\begin{sc}
			\renewcommand{\multirowsetup}{\centering}
			\setlength{\tabcolsep}{5pt}
			\scalebox{1}{
			\begin{tabular}{c|cccc|c}
				\toprule
			    Resolution & FNO \citeyearpar{li2021fourier} & MWT \citeyearpar{Gupta2021MultiwaveletbasedOL} & U-NO \citeyearpar{rahman2022u} & F-FNO \citeyearpar{anonymous2023factorized} & \textbf{LSM (ours)}\\
			    \midrule
                $256$ & 0.00332 & 0.00199 & 0.00450 & 0.00414 & \textbf{0.00123} \\
                $512$ & 0.00333 & 0.00185 & 0.00488 & 0.00347 & \textbf{0.00124} \\
                $1024$ & 0.00377 & 0.00185 & 0.00508 & 0.00319 & \textbf{0.00126} \\
                $2048$ & 0.00346 & 0.00186 & 0.00574 & 0.00313 & \textbf{0.00115} \\
                $4096$ & 0.00324 & 0.00185 & 0.00571 & 0.00314 & \textbf{0.00122} \\
                $8192$ & 0.00336 & 0.00178 & 0.00575 & 0.00315 & \textbf{0.00105} \\
				\bottomrule
			\end{tabular}}
		\end{sc}
	\end{small}
	\vspace{-5pt}
\end{table}

\section{Hyperparameter Sensitivity}\label{appendix:hyperparameter}
As shown in Table \ref{tab:hyperparam}, we test the hyperparameter sensitivity of our model by changing one hyperparameter and fixing the other. Here are the details: 
\begin{itemize}
    \item Change the number of latent tokens $C$ and fix $N=24, K=5$. we can find that the performance of LSM is stable w.r.t.~different choices of $C$, which may come from the equivalence of different latent tokens.
    \item Change the number of basis operators $N$ and fix $C=4, K=5$. Generally, larger $N$ will bring better results, while larger $N$ will also cause more computation cost and optimization problems, which explains why the model performance drops slightly at $N=40$ and $N=48$. Note that the $N=0$ setting is equivalent to the without neural spectral block situation, where the model performance will drop seriously.
    \item Change the number of scales $K$ and fix $C=4, N=24$. In general, adding scales $K$ will improve the model's performance. But the model with too many scales is unimplementable due to the limitation of input resolution.
\end{itemize}
\vspace{-5pt}
Overall, LSM is stable to these three hyperparameters, where $C$ is robust and easy to tune in the range of $3$ to $5$, $N$ is robust in $24$ to $40$ and $K$ is stable in $5$ to $7$. Thus, the setting of number of latent tokens $C$ as $4$, number of basis operators $N$ as $24$ and number of scales $K$ as $5$ can aptly trade off the efficiency and performance.

\begin{table}[tbp]
    \caption{Model performances on Elasticity-G with different number of latent tokens $C$, number of basis operators $N$ and number of scales $K$. ``/'' means that the experiment is unimplementable.}
	\label{tab:hyperparam}
	\vskip 0.1in
	\centering
	\begin{small}
		\begin{sc}
			\renewcommand{\multirowsetup}{\centering}
			\setlength{\tabcolsep}{8pt}
			\scalebox{1}{
			\begin{tabular}{l|ccccccc}
				\toprule
			    Number of Latent Tokens $C$ & 0 & 1 & 2 & 3 & 4 & 5 & 6 \\
                \midrule
                MSE & / & 0.0415 & 0.0415 & 0.0409 & 0.0408 & 0.0411 & 0.0415 \\
                \toprule
			    Number of Basis Operators $N$ & 0 & 8 & 16 & 24 & 32 & 40 & 48 \\
                \midrule
                MSE & 0.0433 & 0.0415 & 0.0418 & 0.0408 & 0.0406 & 0.0413 & 0.0416\\
                \toprule
                    Number of Scales $K$ & 3 & 4 & 5 & 6 & 7 & 8 & 9 \\
                \midrule
                MSE & 0.0428 & 0.0412 & 0.0408 & 0.0400 & 0.0402 & / & / \\
				\bottomrule
			\end{tabular}}
		\end{sc}
	\end{small}
	\vspace{-10pt}
\end{table}

\section{Training Stability}\label{appendix:training}
\begin{figure*}[h]
\begin{center}
\centerline{\includegraphics[width=1.0\textwidth]{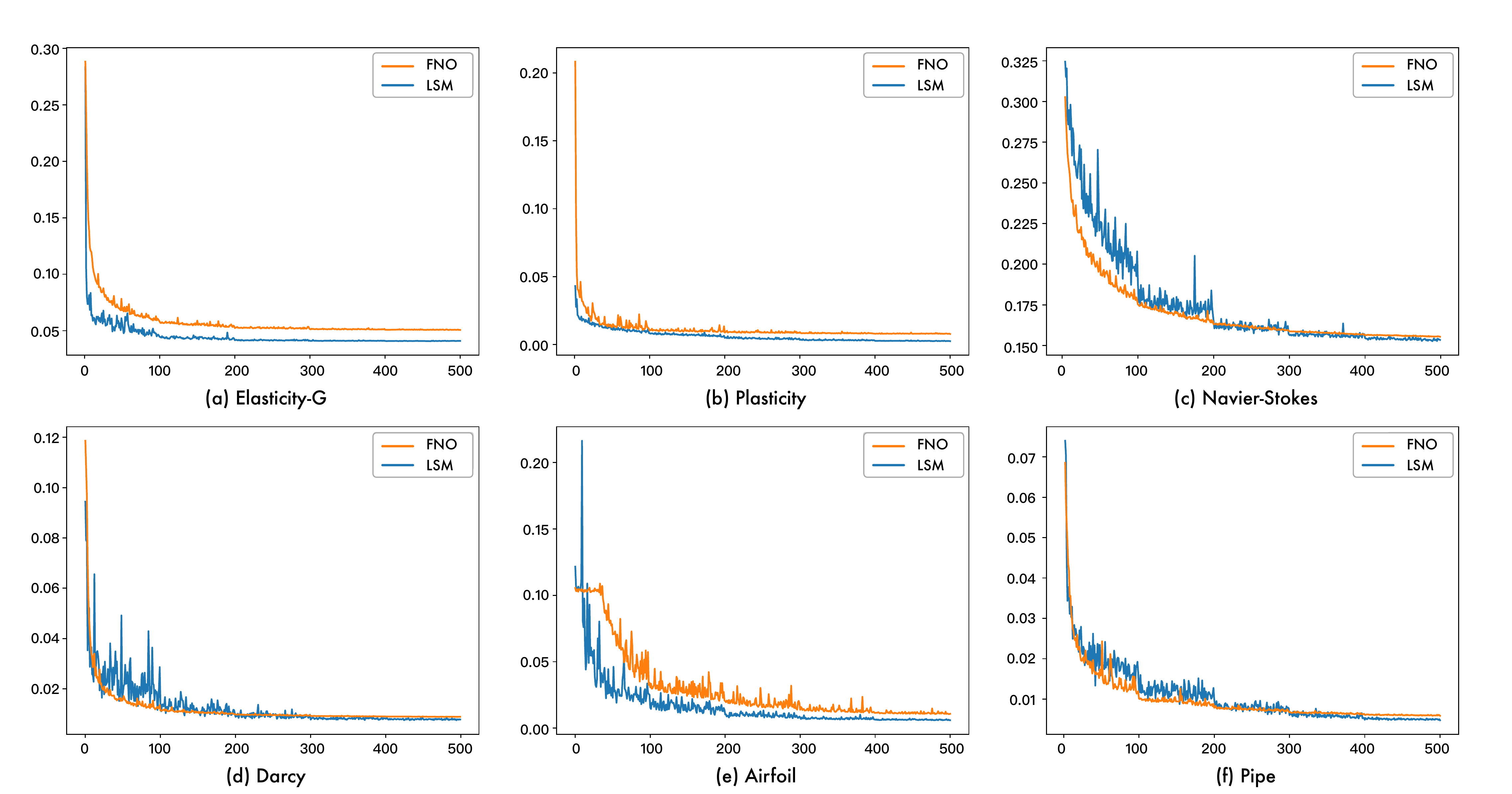}}
    \vspace{-5pt}
	\caption{Model training curves, where the x-axis means the number of epochs and the y-axis is MSE performance in the test set.}
	\label{fig:training_curves}
\end{center}
\vspace{-20pt}
\end{figure*}

We provide the model training curves on different benchmarks in Figure \ref{fig:training_curves}. From Figure \ref{fig:training_curves}, we can observe that in addition to the consistent state-of-the-art performance in all benchmarks, LSM also presents comparable training stability w.r.t.~the well-acknowledged FNO \cite{li2021fourier}. 

Besides, we also repeat all the experiments five times, where the standard deviations of LSM performance are within 0.0001 for Elasticity-P, Elasticity-G and Plasticity, Darcy and Airfoil, and within 0.0002 for Navier-Stokes and Pipe.

\section{Implementation Details} \label{appendix:details}
All experiments are repeated five times, implemented in PyTorch \cite{Paszke2019PyTorchAI} and conducted on a single NVIDIA RTX 3090 24GB GPU. We have provided the training curves and standard deviations in Appendix \ref{appendix:training}. For all methods, the performance at the final epoch is recorded as the final result. Here are the implementation details of the LSM model.

\subsection{Model Configurations}
Here, we present the detailed model configurations for LSM. In the beginning, we will pad the input with zeros properly to resolve the division problem in model configurations.

\begin{table}[h]
    \vspace{-12pt}
	\caption{Model configurations for LSM.}
	\label{tab:model_config}
	\vskip 0.1in
	\centering
	\begin{small}
		\begin{sc}
			\renewcommand{\multirowsetup}{\centering}
			\setlength{\tabcolsep}{4pt}
			\scalebox{1}{
			\begin{tabular}{l|c|c}
				\toprule
			    Model Designs & Hyperparameters & Values \\
			    \midrule
			& Number of latent tokens $C$ & $4$ \\
			& Number of scales $K$ & $5$ \\
			Hierarchical Projection & Downsample Ratio $r=\frac{|\mathcal{D}^{k+1}|}{|\mathcal{D}^{k}|}$ & $0.5$ \\
			Network & Channels of each scale $\{d_{\text{model}}^{1},\cdots,d_{\text{model}}^{K}\}$ & $\{32, 64, 128, 128, 128\}$ \\
			& Channels of latent tokens at each scale $\{d_{\text{latent}}^{1},\cdots,d_{\text{latent}}^{K}\}$ & $\{32, 64, 128, 128, 128\}$ \\
			& Patches of each scale $\{P_{1},\cdots,P_{K}\}$ & $\{256, 64, 16, 4, 1\}$ \\
                  \midrule
			\multirow{1}{*}{Neural Spectral Block} & Number of basis operators $N$ & 24 \\
				\bottomrule
			\end{tabular}}
		\end{sc}
	\end{small}
	\vspace{-10pt}
\end{table}

\subsection{Model Architecture}
In this section, we will illustrate the operations in the patchified multiscale architecture.
\paragraph{Downsample.} Given deep features $\{\boldsymbol{x}^{k}(\mathbf{s})\}_{\mathbf{s}\in\mathcal{D}^{k}}$ at the $k$-th scale, the downsample operation is to aggregate deep features in a local region with maximum pooling and convolution operations, which can be formulized as follows:
\begin{equation}
    \{\boldsymbol{x}^{k+1}(\mathbf{s})\}_{\mathbf{s}\in\mathcal{D}^{k+1}} = \operatorname{Conv}\Big(\operatorname{MaxPool}\big(\{\boldsymbol{x}^{k}(\mathbf{s})\}_{\mathbf{s}\in\mathcal{D}^{k}}\big)\Big),\ \text{$k$ from $1$ to $(K-1)$}.
\end{equation}
\paragraph{Upsample.} Given the deep features $\{\boldsymbol{{\widehat{y}}}^{k+1}(\mathbf{s})\}_{\mathbf{s}\in\mathcal{D}^{k+1}}$, $\{\boldsymbol{{\widehat{y}}}^{k}(\mathbf{s})\}_{\mathbf{s}\in\mathcal{D}^{k}}$ at the $(k+1)$-th and $k$-th scales respectively, which have been projected from latent space back to coordinate space, the upsample process is to fuse the interpolated $k+1$-th features and the $k$-th features with local convolution, which can be formulized as follows:
\begin{equation}
    \{\boldsymbol{{\widehat{y}}}^{k}(\mathbf{s})\}_{\mathbf{s}\in\mathcal{D}^{k}} = \operatorname{Conv}\Bigg(\operatorname{Concat}\Big(\Big[\operatorname{Interpolation}\big(\{\boldsymbol{{\widehat{y}}}^{k+1}(\mathbf{s})\}_{\mathbf{s}\in\mathcal{D}^{k+1}}\big), \{\boldsymbol{{\widehat{y}}}^{k}(\mathbf{s})\}_{\mathbf{s}\in\mathcal{D}^{k}}\Big]\Big)\Bigg),\ \text{$k$ from $(K-1)$ to $1$},
\end{equation}
where we adopt the bilinear $\operatorname{Interpolation}(\cdot)$ for 2D data and the trilinear $\operatorname{Interpolation}(\cdot)$ for 3D data.
\paragraph{Patchify and De-Patchify.} The patchify operation is to split the coordinate set into several non-overlapping local regions with an equal number of coordinates. This process of patchify at the $k$-th scale is formulized as follows:
\begin{equation}
    \{\mathcal{D}^{k}_{j}\}_{j=1}^{P_{k}}=\operatorname{Patchify}(\mathcal{D}^{k}).
\end{equation}
And the depacthify operation is just to splice the patches in different local regions, that is $\mathcal{D}^{k}=\operatorname{De\text{-}Patchify}(\{\mathcal{D}^{k}_{j}\}_{j=1}^{P_{k}})$.

\subsection{Benchmark Construction}
\paragraph{Time-dependent tasks.} In our benchmarks, both Plasticity and Navier-Stokes are time-dependent. For the Plasticity benchmark, since its input is the boundary condition and output is the mesh displacement over time, we adopt the 3D-version LSM for Plasticity experiments, where all the convolution ($\operatorname{Conv}$), max pooling ($\operatorname{MaxPool}$), interpolation ($\operatorname{Interpolation}$) and patchify ($\operatorname{Patchify}$) operations are in the 3D space. As for the Navier-Stokes, since it is an autoregressive task, we still adopt the 2D-version LSM like other benchmarks and predict the next frame step by step. Note that the neural spectral block is applied to the independent latent tokens, hence it is unchanged for both 2D- and 3D-versions. 

\paragraph{Baselines.} We implement all the baselines based on their official code. Note that we focus on the operator-learning paradigm, thus we only adopt their model and uniformly use the L2 loss during training for fairness. Especially, for the MWT \cite{Gupta2021MultiwaveletbasedOL}, we pad the inputs with zeros to make the input resolutions as integer power of two.

\section{Efficiency} \label{appendix:efficiency}
To present a clear efficiency comparison among different models, we fix the model input as a regular grid with the resolution of $S\times S$, the input channel as $1$ and the batch size as $1$. Then, we record the model parameter, GPU memory and running time under different choices of $S$, which are selected from $\{64, 128, 256, 512, 1024\}$. Besides, we also calculate the performance ranking of baselines on each benchmark and present the model efficiency in the order of averaged ranking.

From Table \ref{tab:all_efficiency}, we can obtain the following observations:
\begin{itemize}
    \item \emph{There is an evident gap between solid and fluid physics}. From Table \ref{tab:all_efficiency}, we can find that the performances of baselines are quite different in solid and fluid physics. Concretely, the top 5 models in solid and fluid benchmarks are distinct, except LSM (rank 1st) and F-FNO (rank 5th).  This result also indicates that there is a large gap between solid and fluid physics and previous methods cannot cover different disciplines of physics well.
    \item \emph{LSM presents favorable generality in varied physics.} From the performance ranking on seven tasks, it is observed that other baselines fluctuate greatly in different benchmarks. In contrast, it is impressive that our proposed LSM can achieve consistent state-of-the-art on these varied physics, demonstrating the model generality.
    \item \emph{LSM presents competitive efficiency in high-dimensional inputs.} We have provided the efficiency comparison for $256\times 256$ inputs in Figure \ref{fig:efficiency} of the main text. If we focus on the running time for the inputs with the resolution of $512\times 512$ and $1024\times 1024$, we can find that LSM is clearly faster than U-NO (rank 2nd), U-Net (rank 3rd).
\end{itemize}

\begin{table*}[tbp]
  \vspace{-5pt}
  \caption{Model efficiency comparison and their rankings in solid, fluid and all seven benchmarks, where we select the top 10 methods. A smaller ranking means better performance. Efficiency is evaluated on inputs with size $S\times S$ during the training phase. The batch size is set to 1. Running time is averaged from $10^3$ iterations. ``/'' indicates the out-of-memory situation. }\label{tab:all_efficiency}
  \vskip 0.05in
  \centering
  \begin{threeparttable}
  \begin{small}
  \renewcommand{\multirowsetup}{\centering}
  \setlength{\tabcolsep}{7pt}
  \begin{tabular}{c|c|ccc|c|c|c|c}
    \toprule
    \multicolumn{2}{c}{Input Size ($S \times S$)} & Parameter & GPU Memory & Running Time & \multicolumn{4}{c}{Ranking}  \\
    \cmidrule{6-9}
    \multicolumn{1}{c}{Model} & \multicolumn{1}{c}{$S$} & (MB) & (MB) & (s~/~iter) & Seven tasks$^{\star}$ & Solid$^\ast$ & Fluid$^\dagger$ & Averaged$^\ddagger$ \\
    \toprule
     & 64& 2.002 & 1409 & 0.0353 &\multirow{5}{*}{(1, 1, 1, 1, 1, 1, 1)} & \multirow{5}{*}{1.0} & \multirow{5}{*}{1.0} & \multirow{5}{*}{1.0}\\
    & 128& 2.002 & 1679 & 0.0359& & & & \\
    \textbf{LSM} & 256 & 2.002 & 1959 & 0.0411 & & & & \\
     \textbf{(ours)}& 512 & 2.002 & 3019 & 0.0602 & & & & \\
     & 1024 & 2.002 & 7859 & 0.2002 & & & & \\
    \midrule
    \multirow{5}{*}{U-NO} & 64 & 1.307 & 1345 & 0.0347& \multirow{5}{*}{(6, 2, 2, 4, 7, 4, 9)} & \multirow{5}{*}{3.3} & \multirow{5}{*}{8.0} & \multirow{5}{*}{4.9} \\
     & 128 & 1.307 & 1381& 0.0354& & & & \\
     & 256 & 1.307 & 1603& 0.0397& & & & \\
     & 512 & 1.307 & 2473& 0.0989& & & & \\
     & 1024 & 1.307& 6833& 0.3335& & & & \\
    \midrule
    \multirow{5}{*}{U-Net} & 64 & 4.332 & 1171& 0.0321& \multirow{5}{*}{(3, 8, 5, 6, 4, 6, 4)} & \multirow{5}{*}{5.3} & \multirow{5}{*}{5.0} &  \multirow{5}{*}{5.1} \\
     & 128 & 4.332 & 1243& 0.0307& & & & \\
     & 256 & 4.332 & 1515& 0.0450& & & & \\
     & 512 & 4.332& 2429& 0.1589& & & & \\
     & 1024 & 4.332& 6235& 0.8100& & & & \\
    \midrule
    \multirow{5}{*}{FNO} & 64 & 2.368 & 1137& 0.0202& \multirow{5}{*}{(2, 6, 6, 3, 6, 8, 5)} & \multirow{5}{*}{4.6} & \multirow{5}{*}{7.3} & \multirow{5}{*}{5.1} \\
     & 128 &2.368 & 1179& 0.0203& & & &  \\
     & 256 &2.368 & 1349& 0.0147& & & &  \\
     & 512 &2.368 & 1975& 0.0401 & & & &  \\
     & 1024 &2.368 & 4591& 0.1270& & & &  \\
    \midrule
    \multirow{5}{*}{HT-Net} & 64 & 3.285 & 1175& 0.0406 & \multirow{5}{*}{(10, 3, 10, 5, 3, 2, 3)} & \multirow{5}{*}{7.6} & \multirow{5}{*}{3.2} & \multirow{5}{*}{5.1} \\
     & 128 &3.285 & 1283& 0.0415& & & & \\
     & 256 & 3.285 & 1749& 0.0469& & & & \\
     & 512 & 3.285& 3267& 0.1300& & & & \\
     & 1024 & 3.285& 9259& 0.4581& & & & \\
    \midrule
    \multirow{5}{*}{F-FNO} & 64 & 0.218& 1089& 0.0303& \multirow{5}{*}{(7, 4, 4, 9, 2, 5, 6)} & \multirow{5}{*}{5.0} & \multirow{5}{*}{5.5} & \multirow{5}{*}{5.3} \\
     & 128 &0.218 &1169 & 0.0303& & & & \\
     & 256 & 0.218 &1437 &0.0202 & & & & \\
     & 512 &0.218 &2457 & 0.0825& & & & \\
     & 1024 & 0.218&6443 & 0.3248 & & & & \\
    \midrule
    \multirow{5}{*}{U-FNO} & 64 & 3.990 & 1169& 0.0400& \multirow{5}{*}{(4, 5, 3, 7, 9, 9, 2)} & \multirow{5}{*}{4.0} & \multirow{5}{*}{6.7} & \multirow{5}{*}{5.6} \\
     & 128 & 3.990 & 1241& 0.0400& & & & \\
     & 256 & 3.990& 1499& 0.0471& & & & \\
     & 512 & 3.990 & 2537& 0.1047& & & & \\
     & 1024 & 3.990& 6869& 0.3217& & & & \\
    \midrule
    \multirow{5}{*}{WMT} & 64 & 3.106 & 1145& 0.0615& \multirow{5}{*}{(9, 7, 7, 2, 5, 3, 7)} & \multirow{5}{*}{7.6} & \multirow{5}{*}{4.2} & \multirow{5}{*}{5.7} \\
     & 128 & 3.106 & 1201& 0.0720& & & & \\
     & 256 & 3.106& 1407& 0.0900& & & & \\
     & 512 & 3.106& 2165& 0.1118& & & & \\
     & 1024 & 3.106& 5241& 0.3120& & & & \\
    \midrule
    \multirow{5}{*}{Galerkin} & 64 & 6.319& 1233& 0.0252& \multirow{5}{*}{(5, 10, 8, 10, 8, 7, 8)} & \multirow{5}{*}{7.6} & \multirow{5}{*}{8.2} & \multirow{5}{*}{8.0} \\
     & 128 & 6.319& 1277& 0.0260& & & & \\
     & 256 & 6.319& 1675& 0.0681& & & & \\
    Trasnformer & 512 & 6.319& 3175& 0.2225& & & & \\
     & 1024 & 2.002 & 9333 & 0.8688 & & & & \\
    \midrule
    \multirow{5}{*}{Swin} & 64 &0.538 &1135 &0.0615 & \multirow{5}{*}{(8, 9, 9, 8, 10, 10, 10)} & \multirow{5}{*}{8.6} & \multirow{5}{*}{9.5} & \multirow{5}{*}{9.1} \\
     & 128 & 0.538 & 1261& 0.0333& & & & \\
     & 256 & 0.538& 1789& 0.0355& & & & \\
    Trasnformer & 512 & 0.538& 3759& 0.1150& & & & \\
     & 1024 & /& /& /& & & & \\
    \bottomrule
    \end{tabular}
    \begin{tablenotes}
        \item[] $\star$ The rankings are presented in the order of (Elasticity-P, Elasticity-G, Plasticity, Navier-Stokes, Darcy, Airfoil, Pipe).
        \item[] $\ast$  Top 5 methods of solid benchmarks: LSM (ours), U-NO \citeyearpar{rahman2022u}, U-FNO \citeyearpar{Wen2021UFNOA}, FNO \citeyearpar{li2021fourier}, F-FNO \citeyearpar{anonymous2023factorized}.
        \item[] $\dagger$  Top 5 methods of fluid benchmarks: LSM (ours), HT-Net \citeyearpar{anonymous2023htnet}, WMT \citeyearpar{Gupta2021MultiwaveletbasedOL}, U-Net \citeyearpar{ronneberger2015u}, F-FNO \citeyearpar{anonymous2023factorized}.
        \item[] $\ddagger$  Top 5 methods of all benchmarks: LSM (ours), U-NO \citeyearpar{rahman2022u}, U-Net \citeyearpar{ronneberger2015u}, FNO \citeyearpar{li2021fourier}, HT-Net \citeyearpar{anonymous2023htnet}.
    \end{tablenotes}
    \end{small}
    \end{threeparttable}
\end{table*}


\section{Limitations}\label{appendix:limitations}
As we discussed in the main results (Table \ref{tab:mainres}), performance under various resolutions (Table \ref{tab:resolution}), efficiency comparison (Table \ref{tab:all_efficiency}) and transfer learning tasks (Table \ref{tab:transferability_all}), LSM can precisely solve the PDEs with good efficiency and transferability, covering both solid and fluid physics and various geometrics. Although LSM can achieve advanced performance, it still holds some limitations. Here are the discussions.


One potential limitation of LSM may lie in the model generality among different PDEs, where we expect a zero-shot PDE solver like the foundation models in natural language processing, such as GPT-3 \cite{NEURIPS2020_1457c0d6}, T5 \cite{raffel2020exploring} and etc. Note that due to the inherent complexity of PDEs, a small perturbation to the coefficients of PDEs may change their property seriously, such as condition number, with or without explicit solution, the convergence of infinite values \cite{evans2010partial}. Thus, to tackle this potential limitation, we need first to explore the fundamental question of ``whether there is a universal solution to all PDEs or not,'' which is clearly far beyond the scope of our paper. Thus, we would like to leave this problem as future work.

\section{Societal Impacts}\label{appendix:impact}
\paragraph{Real-world applications.} In this paper, we present the LSM as a practical deep solver for high-dimensional PDEs. Given the state-of-the-art performance of LSM, this paper may help many PDE-related applications, such as airfoil design, the load-bearing tests of civil engineering, weather forecasting, etc. Especially, LSM can also present a favorable transferability to limited data scenarios (Table \ref{tab:transferability_all}), which is important for fast-adaption to new scenarios.

\paragraph{Academic research.} Unlike previous methods, LSM attempts to solve high-dimensional PDEs through a new technology roadmap: going beyond high-dimensional coordinate information and solving PDEs in latent space, which can be a good supplement for the operator learning community.

This paper only focuses on the scientific problem. All the datasets are generated by public tools and strictly follow the corresponding licenses (Appendix \ref{appendix:dataset}). Thus, there is no potential ethical risk or negative social impact.

\end{document}